\documentclass[table]{ai2style/ai2}

\usepackage{microtype}
\usepackage{hyperref}
\usepackage{url}
\usepackage{booktabs, tabularx} 
\usepackage{graphicx}
\usepackage{lineno}
\usepackage{enumitem}
\usepackage{listings} 
\usepackage{svg}

\definecolor{ darkblue}{rgb}{0, 0, 0.5}
\hypersetup{colorlinks=true, citecolor=darkblue, linkcolor=darkblue, urlcolor=darkblue}

\usepackage{amssymb}
\usepackage{multirow}
\usepackage{bigdelim}
\usepackage{todonotes}
\usepackage{longtable}
\usepackage{tabularray}
\usepackage{wrapfig}
\usepackage[most]{tcolorbox}
\usepackage{url}
\usepackage{xspace}
\usepackage{svg}
\usepackage[absolute]{textpos} 

\usepackage{fdsymbol}   

\usepackage[utf8]{inputenc} 
\usepackage[T1]{fontenc}    
\usepackage{url}            
\usepackage{booktabs}       
\usepackage{amsfonts}       
\usepackage{nicefrac}       
\usepackage{microtype}      
\usepackage[table,xcdraw]{xcolor}         
\usepackage{amsmath}
\usepackage[most]{tcolorbox}
\usepackage{csquotes}

\usepackage{siunitx}
\usepackage{graphicx}
\usepackage{arydshln}
\usepackage{wrapfig}
\usepackage{enumitem}
\usepackage{soul} 

\usepackage{multirow}
\usepackage{xspace}
\usepackage{adjustbox}
\usepackage{pifont}
\usepackage{caption}
\usepackage{makecell}
\usepackage{subcaption}
\usepackage{bold-extra}
\usepackage{wrapfig}
\usepackage{url}

\usepackage{pgf-pie}

\usepackage{hyperref}
\definecolor{linkcolor}{RGB}{0, 0, 128}
\hypersetup{
     colorlinks   = true,
     citecolor    = linkcolor,
     linkcolor    = linkcolor,
     urlcolor     = linkcolor,
}
\usepackage{pifont}
\usepackage{listings}

\setlist[itemize]{leftmargin=*,itemsep=0em,parsep=0.3em,topsep=0.3em}

\DeclareUnicodeCharacter{2212}{\ensuremath{-}}

\definecolor{maroon}{HTML}{F26035}
\definecolor{yellow}{HTML}{FDBC42}
\definecolor{lavender}{HTML}{734f96}
\definecolor{darkergrey}{HTML}{444444}
\definecolor{midgrey}{HTML}{e6eded}
\definecolor{ai2pink}{HTML}{f0529c}
\definecolor{ai2midpink}{HTML}{fad3e5}
\definecolor{ai2lightpink}{HTML}{fbecf3}
\definecolor{ai2midwhite}{HTML}{f2e5d9}
\definecolor{ai2offwhite}{HTML}{fbf4ee}
\definecolor{ai2green}{HTML}{0fcb8c}
\definecolor{ai2lightgreen}{HTML}{e7f9f3}
\definecolor{ai2darkgreen}{HTML}{105257}
\definecolor{ai2purple}{HTML}{B932EB}
\definecolor{ai2lightpurple}{HTML}{f7e8fc}
\definecolor{neutralEight}{HTML}{343434}
\definecolor{neutralFive}{HTML}{838383}
\definecolor{neutralThree}{HTML}{bebebe}
\definecolor{neutralOne}{HTML}{dedede}
\definecolor{lightgrey}{HTML}{fafcfc}
\definecolor{plum}{rgb}{0.56,0.27,0.52}

\usepackage{tikz}

\definecolor{maroon}{HTML}{F26035}
\definecolor{yellow}{HTML}{FDBC42}
\definecolor{darkred}{RGB}{156, 39, 33}
\definecolor{darkblue}{RGB}{31, 90, 153}
\definecolor{forestgreen}{rgb}{0.13, 0.55, 0.13}
\definecolor{brickred}{rgb}{0.8, 0.25, 0.33}
\definecolor{olmoDarkBlue}{HTML}{012e59}
\definecolor{olmoBlue}{HTML}{265ed4}
\definecolor{olmoLightBlue}{HTML}{012e59}
\definecolor{olmoTeal}{HTML}{00d5ff}
\definecolor{olmoYellow}{HTML}{ffbb00}
\definecolor{olmoOrange}{HTML}{ff9100}

\usepackage{setspace}

\usepackage{nicematrix}
\newcolumntype{L}[1]{>{\raggedright\let\newline\\\arraybackslash\hspace{0pt}}m{#1}}
\newcolumntype{C}[1]{>{\centering\let\newline\\\arraybackslash\hspace{0pt}}m{#1}}
\newcolumntype{R}[1]{>{\raggedleft\let\newline\\\arraybackslash\hspace{0pt}}m{#1}}
\newcolumntype{P}[1]{>{\centering\let\newline\\\arraybackslash\columncolor{ai2lightpink}}m{#1}}
\newcommand{\allenAiAff}{\raisebox{.28em}{\hspace{.02em}\scalebox{0.7}{\textbf{1}}}}

\newcommand{\uwAff}{\raisebox{.28em}{\hspace{.02em}\scalebox{0.7}{\textbf{2}}}}

\newcommand{\huggingface}{\raisebox{-1.5pt}{\includegraphics[height=1.05em]{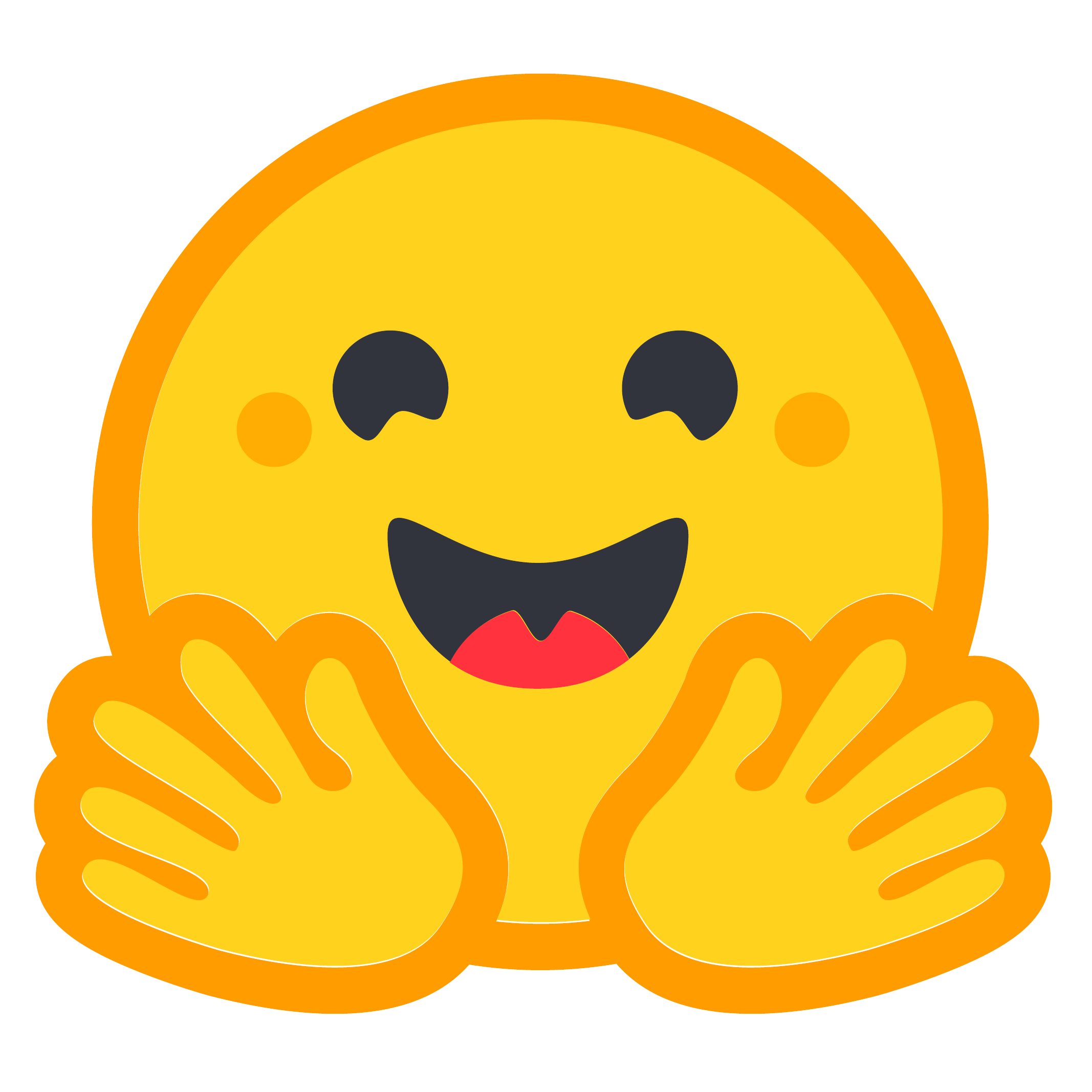}}\xspace}
\newcommand{\hfdataset}{\raisebox{-1.5pt}{\includegraphics[height=1.05em]{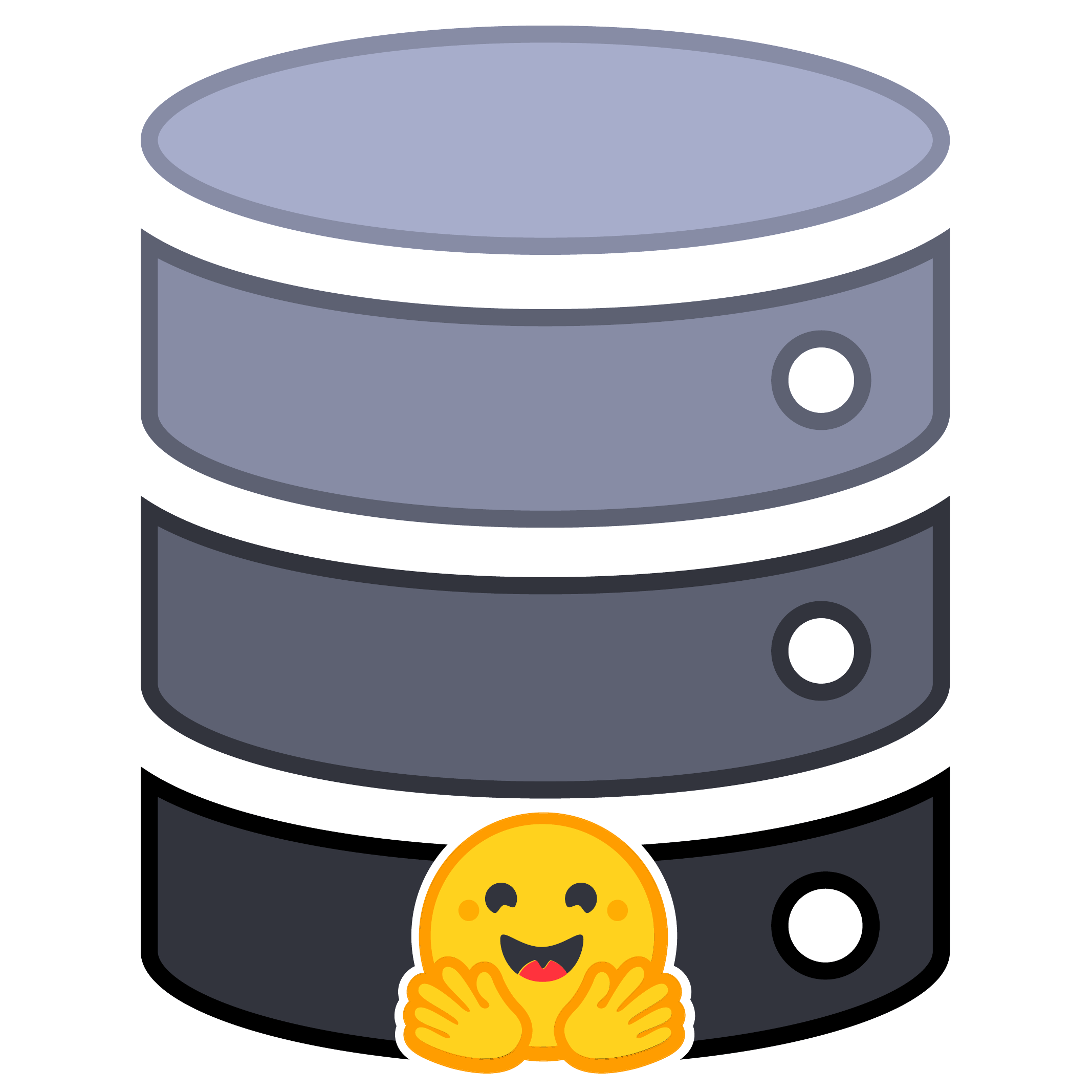}}\xspace}
\newcommand{\emailLogo}{\raisebox{-1.5pt}{\includegraphics[height=1.05em]{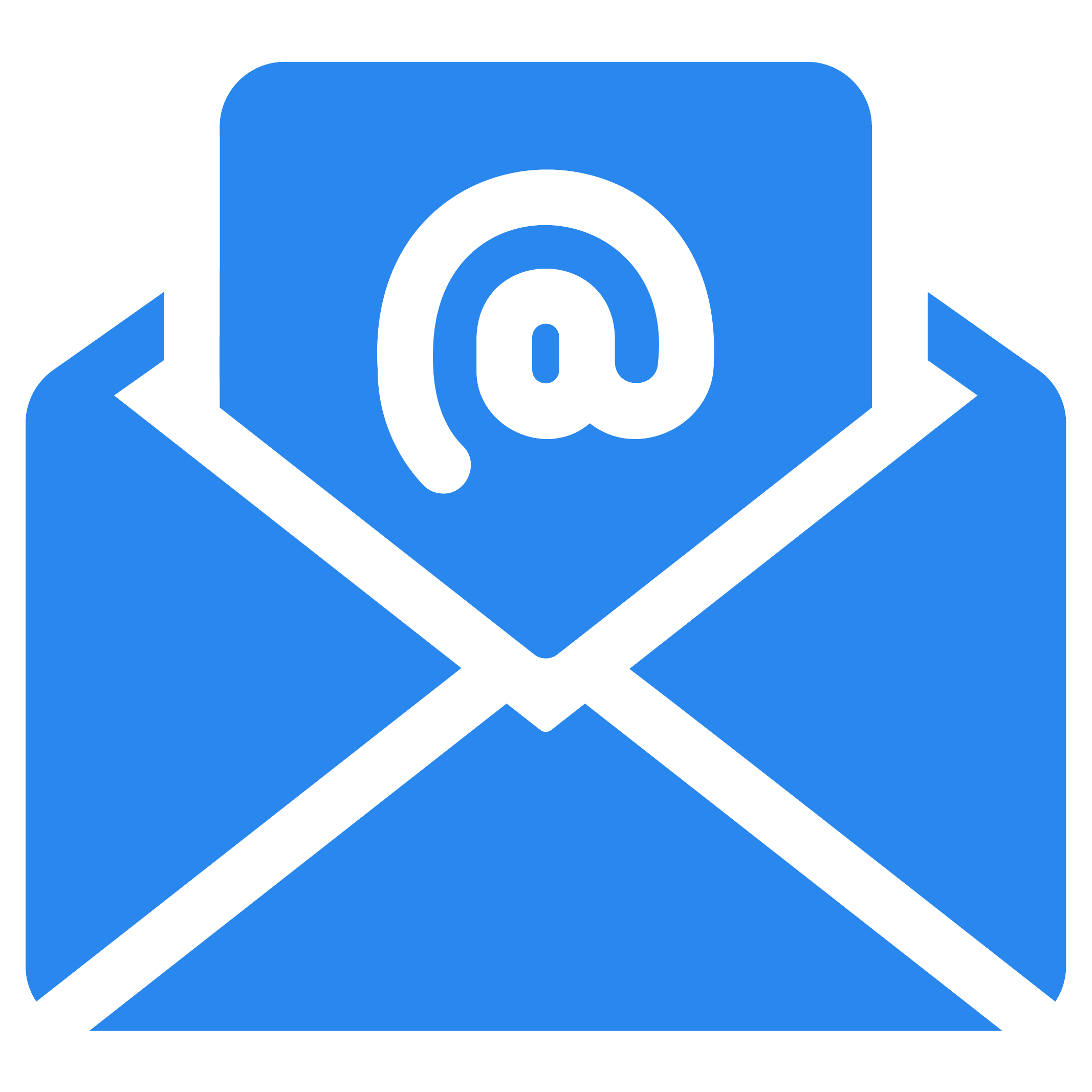}}\xspace}
\newcommand{\github}{\raisebox{-1.5pt}{\includegraphics[height=1.05em]{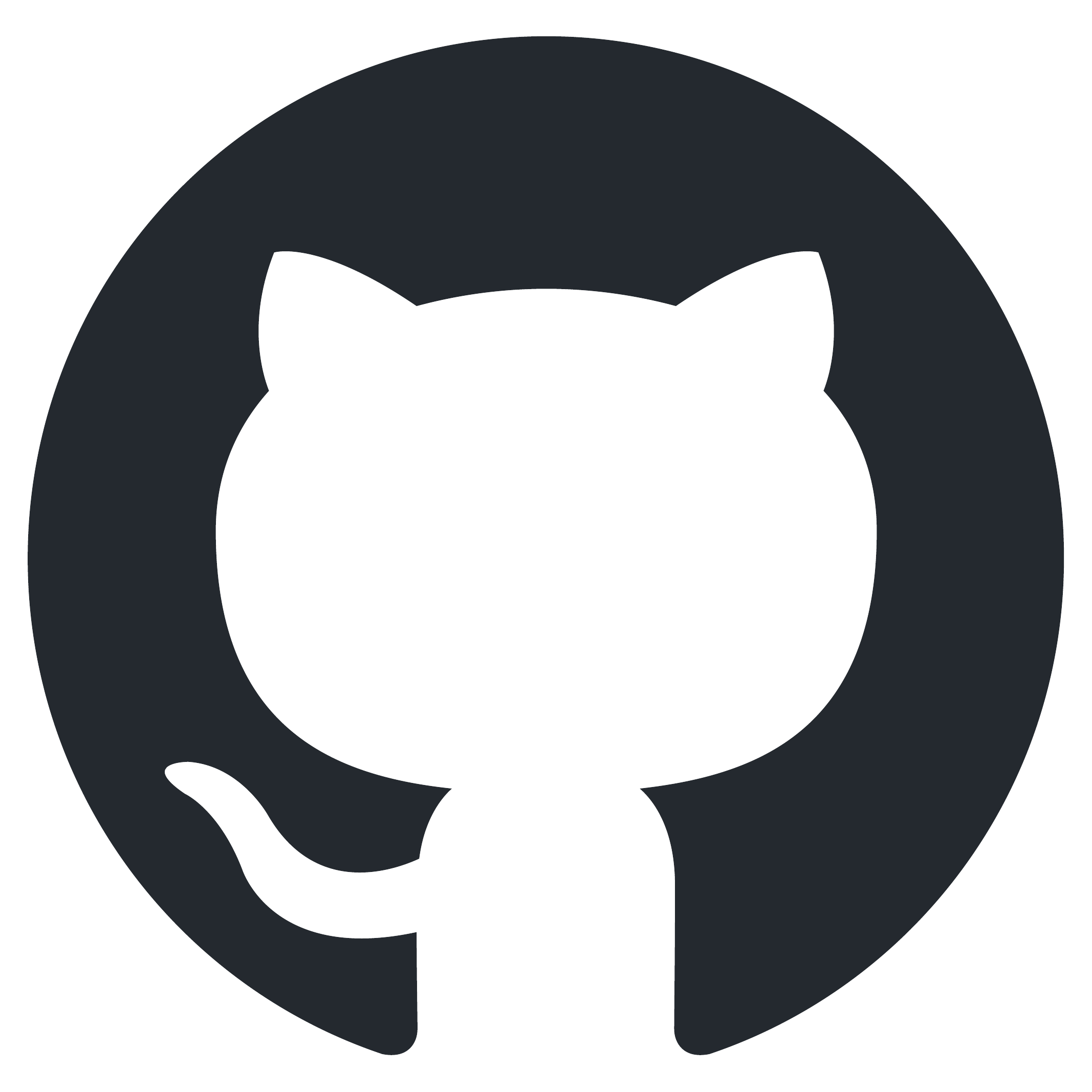}}\xspace}

\newcommand{\patchtoken}{\texttt{<PATCH>} token\xspace}
\newcommand{\patchtokens}{\texttt{<PATCH>} tokens\xspace}
\newcommand{\subpatchtoken}{\texttt{<SUBPATCH>} token\xspace}

\newcommand{\locationtoken}{\texttt{<LOCATION>} token\xspace}

\newcommand{\imagetoken}{\texttt{<IMAGE>} token\xspace}
\newcommand{\imagetokens}{\texttt{<IMAGE>} tokens\xspace}
\newcommand{\guidata}{MolmoPoint-GUISyn}
\newcommand{\trackdata}{MolmoPoint-Track}
\newcommand{\trackreal}{MolmoPoint-TrackAny}
\newcommand{\tracksynth}{MolmoPoint-TrackSyn}

\newcommand{\model}{MolmoPoint-8B\xspace}
\newcommand{\modelImg}{MolmoPoint-GUI-8B\xspace}
\newcommand{\modelVid}{MolmoPoint-Vid-8B\xspace}

\newcommand{\apionlyheader}[1]{%
  \multicolumn{#1}{@{}l}{\textbf{\textit{API call only}}}
}
\newcommand{\openweightsheader}[1]{%
  \multicolumn{#1}{@{}l}{\textbf{\textit{Open weights}}}
}
\newcommand{\fullyopenheader}[1]{%
  \multicolumn{#1}{@{}l}{\textbf{\textit{Fully open}}}
}
\newcommand{\modelheader}[1]{%
  \multicolumn{#1}{@{}l}{\textbf{\textit{MolmoPoint}}}
}

\newcommand{\core}{\textsuperscript{\textcolor{ai2pink}{\ding{170}}}}

\definecolor{molmocolor}{RGB}{240, 82, 156}
\definecolor{tablegray}{RGB}{223, 242, 252}
\definecolor{tablegreen}{RGB}{15, 203, 150}
\definecolor{tableyellow}{RGB}{250, 242, 233}
\definecolor{tableblue}{RGB}{240, 82, 156}
\definecolor{darkpink}{RGB}{139, 14, 98}
\definecolor{baselinecolor}{gray}{.9}

\title{MolmoPoint\\{\fontsize{18pt}{12pt}\selectfont  Better Pointing for VLMs with Grounding Tokens}}

\authorTwo[1]{Christopher Clark\core}
\authorTwo[1]{Yue Yang\core}
\authorTwo[1]{Jae Sung Park\core}
\authorTwo[1,2]{Zixian Ma}
\authorTwo[1,2]{Jieyu Zhang}
\authorTwo[1]{Rohun Tripathi}
\authorTwo[1,2]{Mohammadreza Salehi}
\authorTwo[1]{Sangho Lee}
\authorTwo[1]{Taira Anderson}
\authorTwo[1]{Winson Han}
\authorTwo[1,2]{Ranjay Krishna\core}

\affiliation[\allenAiAff]{Allen Institute for AI}
\affiliation[\uwAff]{University of Washington}

\contribution{\textcolor{ai2pink}{\ding{170}} marks core contributors}

\abstract{
Grounding has become a fundamental capability of vision-language models (VLMs). 
Most existing VLMs point by generating coordinates as part of their text output, which requires learning a complicated coordinate system and results in a high token count.
Instead, we propose a more intuitive pointing mechanism that directly selects the visual tokens that contain the target concept.
Our model generates a special pointing token that cross-attends to the input image or video tokens and selects the appropriate one.
To make this model more fine-grained, we follow these pointing tokens with an additional special token that selects a fine-grained subpatch within the initially selected region, and then a third token that specifies a location within that subpatch.
We further show that performance improves by generating points sequentially in a consistent order, encoding the relative position of the previously selected point, and including a special \textit{no-more-points} class when selecting visual tokens.
Using this method, we set a new state-of-the-art on image pointing (70.7\% on PointBench), set a new state-of-the-art among fully open models on GUI pointing (61.1\% on ScreenSpotPro), and improve video pointing (59.1\% human preference win rate vs. a text coordinate baseline) and tracking (+6.3\% gain on Molmo2Track).
We additionally show that our method achieves much higher sample efficiency and discuss the qualitative differences that emerge from this design change.
}

\metadata[\quad\huggingface Models:]{
\href{https://huggingface.co/allenai/MolmoPoint-8B}{\texttt{\model}} \quad
\href{https://huggingface.co/allenai/MolmoPoint-GUI-8B}{\texttt{\modelImg}} \quad
\href{https://huggingface.co/allenai/MolmoPoint-Vid-4B}{\texttt{\modelVid}}}

\metadata[\vspace{.5em}\quad\hfdataset Data:]{
    \href{https://huggingface.co/datasets/allenai/MolmoPoint-GUISyn}{\texttt{\guidata}} \quad
    \href{https://huggingface.co/datasets/allenai/MolmoPoint-TrackAny}{\texttt{\trackreal}} \quad
    \href{https://huggingface.co/datasets/allenai/MolmoPoint-TrackSyn}{\texttt{\tracksynth}}
}

\metadata[\vspace{.5em}\quad\github Code:]{
    \href{https://github.com/allenai/molmo2}{\texttt{https://github.com/allenai/molmo2}} \quad
    \href{https://github.com/allenai/MolmoPoint-GUISyn}{\texttt{https://github.com/allenai/MolmoPoint-GUISyn}}
}

\metadata[\vspace{.5em} \quad\huggingface Demos:]{\href{https://huggingface.co/spaces/allenai/MolmoPoint-8B-Demo?}{\texttt{\model{}}}
\quad
\href{https://huggingface.co/spaces/allenai/MolmoPoint-GUI-8B-Demo?}{\texttt{\modelImg{}}}}

\metadata[\vspace{.5em}\quad\emailLogo Contact:]{\color{ai2accent}{\texttt{molmo@allenai.org}}}

\begin{document}

\maketitle

\section{Introduction}

\begin{figure}[t]
    \centering
    \includegraphics[width=0.99\linewidth]{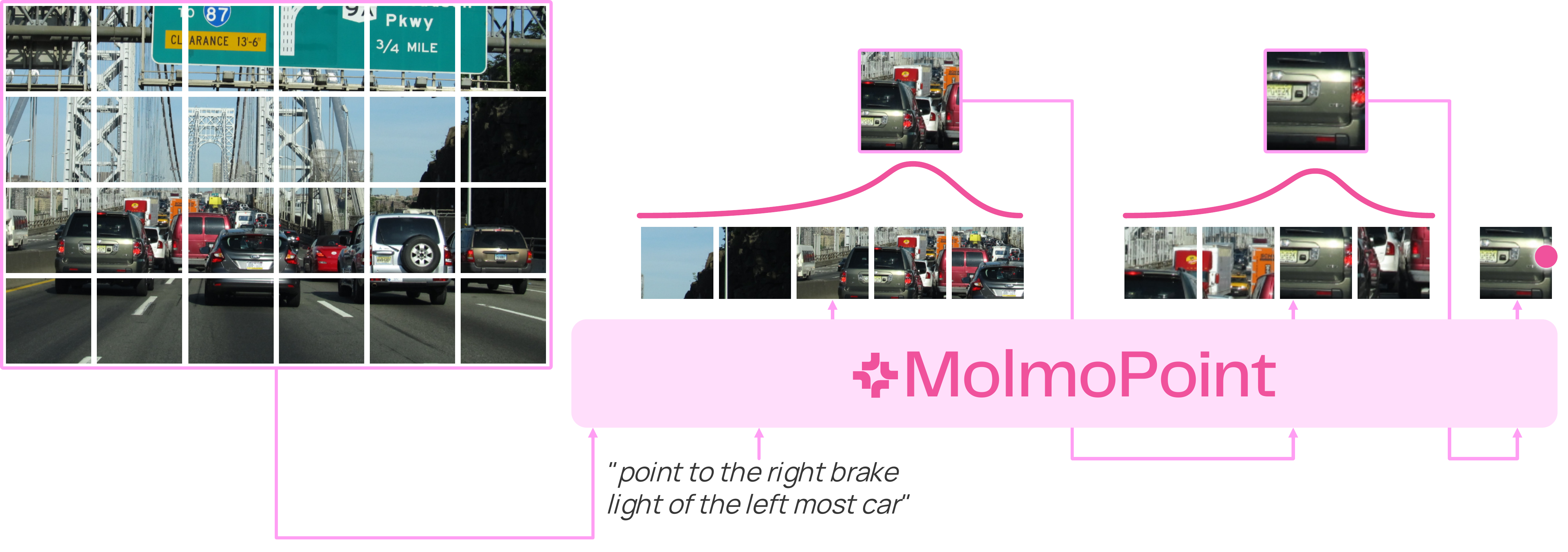}
    \caption{\textbf{Overview of MolmoPoint}. To point, our model scores coarse-grained image patches using the LLM's hidden states, then scores fine-grained subpatches from the highest scoring patch using ViT image features, and then selects a point within the highest scoring subpatch.}
    \label{fig:model}
\end{figure}
Grounding through pointing is a key capability for vision-language models (VLMs). Pointing has direct applications to robotics, where points have been shown to be an effective way for VLMs to build plans for grasping or navigation~\cite{lee2025molmoact,sun2025emma,yang2025bridging}. Computer user agents have increasingly used pointing to determine how to interact with graphical user interfaces (GUIs)~\cite{venusteam2026uivenus15technicalreport,zhou2025maiuitechnicalreportrealworld,qin2025ui,wang2025opencua}. Pointing can also be used with chain-of-thought to improve performance on tasks like counting~\cite{molmov1,molmo2}, and it can be used to refer back to visual input when communicating with users to provide clearer and more interpretable responses.

VLMs typically point in one of two ways: by directly generating text coordinates~\cite{molmov1,qwen3technicalreport,gemini3}, or by generating special tokens that correspond to discretized coordinate bins~\cite{lu2024unified,chen2021pix2seq}. Instead, as shown in Figure ~\ref{fig:model}, we propose to use \textit{grounding tokens} that directly select visual tokens from the input video or image.

To predict a point, the model emits three special grounding tokens, \texttt{<PATCH>}, \texttt{<SUBPATCH>}, and \texttt{<LOCATION>}, that generate a point in a coarse-to-fine manner.
The \patchtoken is used to select a coarse-grained patch in the input image (or video) by attending to the hidden states of the LLM's visual tokens.
The \subpatchtoken selects a subpatch within that patch by attending to the ViT features of finer-grained patches within that patch.
Finally, the \locationtoken selects a point within the subpatch.
When used as input, the \patchtoken and \subpatchtoken use embeddings derived from the selected patch and subpatch.
This allows the model to carry forward location information as it generates future tokens.

To give the model additional awareness of what it has already pointed to, we apply rotary embeddings (RoPE)~\cite{su2024roformer} when selecting a patch to encode how far candidate patches are from the patch selected by the \textit{previous} \patchtoken.
This encoding makes it easier for the model to generate consistent, ordered points and avoid double-pointing. 
We also allow \patchtokens to emit a no-more-points class instead of selecting a token to indicate that the model should stop pointing. We show that this prevents degenerate behavior where the model generates an excessive number of points. 

Our approach has several practical advantages. 
First, the model no longer needs to learn or memorize a coordinate system, which we show makes learning faster and improves generalization across image resolutions unseen during training. 
Second, it reduces the number of output tokens required to represent each point, lowering the decoding cost and improving inference latency. 
Third, it more tightly couples visual recognition and pointing: if the model has already encoded an object, action, or part in the hidden state of a visual token, it becomes trivial to point to that content by generating a query vector that matches its embedding.
We show that this leads to stronger pointing performance and shows signs of improving transfer to tasks beyond grounding.

To explore this approach, we train three models: (1) \textbf{\model{}}: a general-purpose image and video VLM following the Molmo2 pipeline, (2) \textbf{\modelImg{}}: a model specialized for GUI pointing, and (3) \textbf{\modelVid{}}: a lighter-weight model specialized for video pointing. To train \modelImg{}, we construct \textbf{\guidata}, a new synthetic dataset of high-resolution GUI grounding examples by extending the code-guided data generation method of CoSyn~\cite{cosyn}. To improve tracking in \model{}, we also contribute \textbf{\trackdata{}}, a dataset of human-annotated and synthetic tracks for broader object and scene coverage.

We evaluate these models across many pointing tasks. For natural images, \model{} sets a new SoTA on PointBench~\cite{pointarena} and PixMo-Points~\cite{molmov1}, beating the previous methods by 2 points and 4 points, respectively. For GUI pointing, we find \modelImg{} achieves over 5 points better on ScreenSpotPro~\cite{li2025screenspot} and 4 points better on OSWorldG~\cite{xie2025scalingcomputerusegroundinguser} compared to a baseline using text coordinates, and is SoTA among models of a similar size that have open data. For video pointing, \model{} shows a several-point gain on counting metrics and better human preference scores compared to Molmo2, despite being trained on the same data, and \modelVid{} further improves these metrics. For video tracking, \model{} reaches 62.5 on $\mathcal{J}\&\mathcal{F}$ vs 56.7 for Molmo2 and shows large gains from both our new data and model design.
We also show that our approach improves training and sample efficiency and has notable qualitative effects on the pointing behavior. We will release our models, code, and data.

\section{Related Work}
\paragraph{Generating Coordinates.}  
Generating text coordinates or discrete tokens for grounding is an old approach for VLMs~\cite{wang2022ofa,chen2021pix2seq,lu2024unified,lu2022unified}. 
Large-scale pointing datasets such as PixMo-Points~\cite{molmov1} have allowed VLMs to handle pointing across a wide range of objects and images~\cite{molmov1}, and many recent VLMs have adopted this capability~\cite{gemini3,qwen3technicalreport,liu2025visionreasoner,yuan2024robopoint,wang2025internvl3,beyer2024paligemma}.
\model{} shows that using grounding tokens can provide a stronger and more efficient way to learn this skill. 

\paragraph{GUI Grounding.}
Many recent works have developed models that use pointing to interact with graphical user interfaces~\cite{wang2025opencua,wu2025gui,lin2024showui}. Existing methods often try to improve performance by enhancing data generation~\cite{qin2025ui,jedi,groundcua,cheng2024seeclick} or by using reinforcement learning~\cite{venusteam2026uivenus15technicalreport,zhou2025maiuitechnicalreportrealworld,Yuan2025EnhancingVG,Tang2025LPOTA}.
Other works have improved GUI grounding through agentic, multi-step strategies such as zooming in and cropping the input screenshot~\cite{zhou2025maiuitechnicalreportrealworld,zhangmanicog}, although this comes at the expense of higher compute costs. Our work shows that improving the point representation can also significantly enhance GUI grounding.

\paragraph{GUI Grounding Datasets.}
Existing GUI grounding datasets have been built both purely synthetically~\cite{cosyn,ariaui,gou2024navigating,wu2024atlas,gou2024navigating} and with humans~\cite{kapoor2024omniact,chai2024amexandroidmultiannotationexpo,groundcua}. Our \guidata{} differs in that it focuses on high-resolution images and greater diversity across operating systems, websites, software, apps, resolutions, and aspect ratios. \guidata{} also provides extremely dense annotations (54 points per image on average), making it very efficient to train on using \textit{message-trees} to group all annotations for an image into a single training sequence~\cite{molmo2}.

\paragraph{Video Grounding.}
Open-vocabulary video grounding is still generally done by specialized models~\cite{yan2024visa,bai2024one,li2025refsam,ahmad2025videomolmo}, with only a few VLMs supporting this capability~\cite{molmo2,gemini3}.
We believe that grounding should not be limited to images, which is partly why we build on top of the Molmo2 models that support video pointing. Our results suggest that token referencing can help in this domain as well.

\paragraph{Grounding Tokens.}
Grounding tokens have been used for tasks such as image segmentation~\cite{beyer2024paligemma,lai2024lisa,bai2024one,rasheed2024glamm} or depth estimation~\cite{molmoact2025,bigverdi2025perception}. These methods typically employ a pre-trained decoder that constructs the grounded output from tokens. In contrast, our method decodes grounding tokens through lightweight projectors on top of the hidden states, removing the need for pre-trained decoders.

More similar to our work, PaDT~\cite{su2025patch} adds tokens to the model's vocabulary using hidden states of input vision tokens, which allows generated tokens to similarly cross-attend to the input visual tokens.
However, their approach uses a separate decoder to obtain bounding boxes or other grounding information from those tokens, whereas our method uses the spatial location of visual tokens, along with refinement with additional tokens, to point. Our method also applies this approach to videos and GUIs.

GUI-Actor~\cite{wu2025gui} also allows cross-attention between a special \texttt{<ACTOR>} token and visual patches; however, it does not add refinement stages to allow high-precision pointing and only applies their methods to GUIs and single points.
\section{Method}
\begin{figure}[!t]
    \centering
    \includegraphics[width=\linewidth]{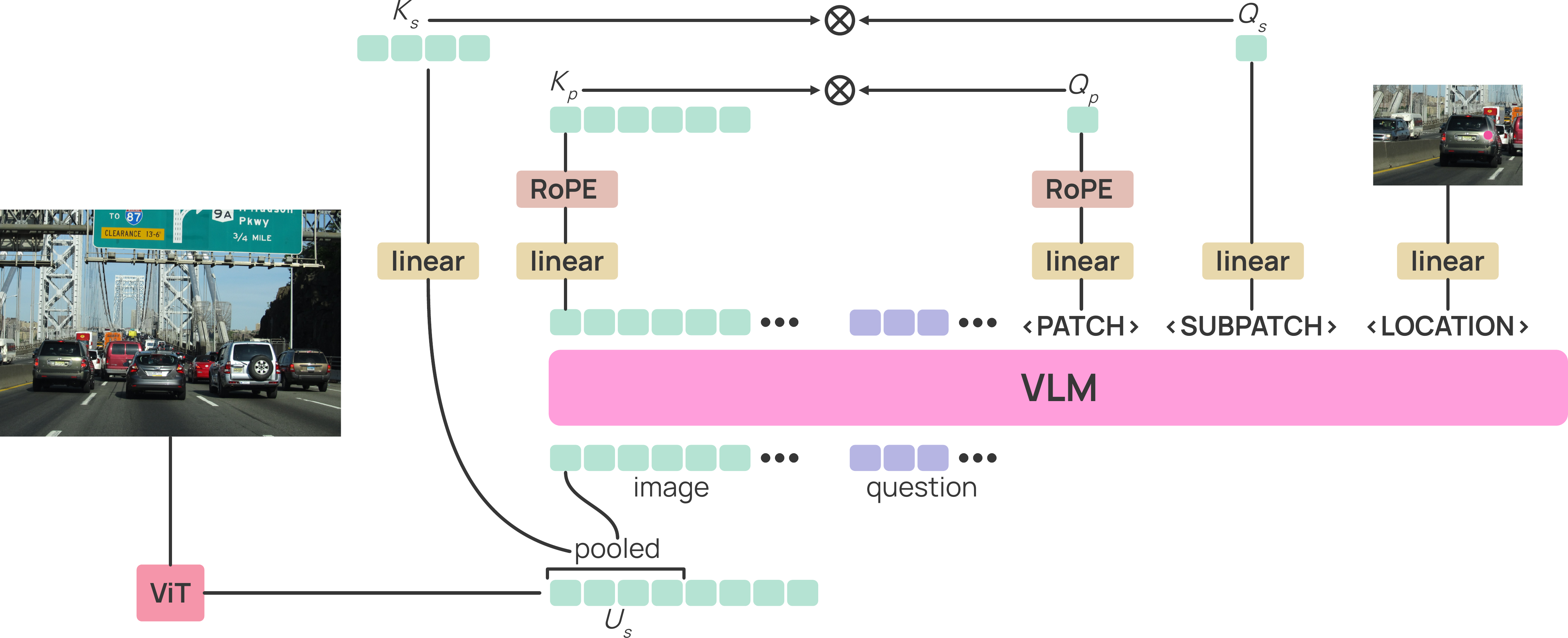}
    \caption{\textbf{Pointing with grounding tokens}. Keys are built from image tokens and ViT patch features, and queries are built from the \patchtoken and \subpatchtoken hidden states, to score patches and subpatches. The \locationtoken predicts the final output points within the highest scoring subpatch.}
    \label{fig:details}
\end{figure}

Our approach trains the model to point by directly selecting which visual token contains the target object and then refining that location by generating additional tokens.
We describe it in more detail below.

\subsection{Patch Selection}
First, we add a special \patchtoken to the model's vocabulary. When this token is generated, a query vector is built from its hidden state: 

$$q_p = W_{pq}\text{Norm}(h_p)$$

Here $W_{pq}$ is a learned parameter with shape $M \times D$, $h_p$ is the hidden state of the \patchtoken with shape $D \times 1$, Norm is a layer norm layer, $D$ is the model's hidden size, and $M$ is a hyper-parameter. We also generate key vectors for each \imagetoken that embeds visual input as: 
$$K_p = W_{pk}\text{Norm}(H_i)$$ 
where $W_{pk}$ is another learned parameter with shape $M \times D$, $H_i$ are the hidden states of the \imagetokens with shape $D \times I$, and $I$ is the number of image tokens. 
Finally, we score each image token as
$$s_p = K_p^\top q_p / \sqrt{M}$$

The score vector $s_p$ has shape $I \times 1$. 
During training, we compute the loss of this selection process as:
$$L_p = \text{cross\_entropy}(\text{softmax}({s_p}), t_p)$$
where $t_p$ is the ground truth target token. 
$L_p$ is added directly to the token-level loss from the LLM before that loss is averaged by the number of tokens.

During inference, we select the highest scoring token $p^* = \text{argmax}(s_p)$. During training, we instead use $p^* = t_p$.
Then, when \patchtoken is an input, we add the input embedding of the \imagetoken that was selected to its embedding: $q_p+E_i[:,p^*]$, where $E_i$ are input embeddings of the \imagetokens. This is important so the model is aware of which token it pointed to. 

During training, we sort ground truth points so that the \imagetokens the \patchtokens select are ordered based on where they appear in the input sequence. We mask out \imagetokens that come before previously selected \imagetokens during both training and inference to enforce this pattern.

\subsection{Location Refinement}
In most VLMs, image tokens are constructed by pooling multiple patches from the underlying ViT. For example, in Molmo2 models, each \imagetoken is built from 4 ViT patches that each cover 14x14 pixels, so it represents a 28x28 pixel area. This is too coarse-grained, so we refine that location by adding additional tokens after the \patchtoken. 

After a \patchtoken, our model also emits a \subpatchtoken that selects one of the ViT patches that were pooled to build $p*$. This is done through dot-product scoring as before. The hidden state of the \subpatchtoken $h_s$ is projected to create a query vector $q_s$, and key vectors $K_s$ are built by projecting the ViT features for the subpatches $U_s$, where $U_s$ is a $T \times K$ matrix, $K$ is the number of subpatches, and $T$ is the dimensionality of the ViT.
 
 We similarly use the ground-truth subpatch location to compute a loss $L_s$ for this component during training and select a subpatch index $s^*$.
When a \subpatchtoken is used as input, its embedding is built from the hidden state of the selected ViT patch: $q_s+W_{se} U_s[:,s^*]$ where $W_{se}$ with shape $D \times T$ projects the ViT patch feature to the LLM's dimension. Adding this embedding indicates to the LLM which subpatch was selected and gives the model access to the unpooled features of the selected patch, which we find important when trying to further refine the location.

This gives us a 14x14 resolution, which can still be too coarse-grained. To produce a precise point, we emit a final \locationtoken. The hidden state of the \locationtoken is used to predict one of 9 locations within the subpatch (arranged in a 3x3 grid) using a single linear layer. With 14x14 ViT patches, this results in a precision of about 4.7 pixels. Unlike pointing with text coordinates, this method maintains a 4-pixel resolution regardless of input size, potentially enabling fine-grained pointing even with ultra-HD images.

\subsection{Rotary Embedding}
We add rotary embeddings to better encode how \imagetokens are positioned relative to the previously selected \patchtoken. This is important to help the model follow the sorted order of points or to track what frames the previous points were generated for when doing video pointing.

This is implemented by rotating the \patchtoken key and query vectors:
$$s_p = \text{Rot}(K_p, p_i)^\top \text{Rot}(q_p, p_q) / \sqrt{M}$$
Where $p_i$ contains the \imagetoken position  $[0, 1, 2, ...., I]$ and $p_q$ is the image position selected by the previous \patchtoken, or 0 if there is no such \patchtoken.

\subsection{No-More-Points Class}
One issue with this approach is that if the model chooses to generate a \patchtoken, it is forced to select a point, even if none of the scores in $s_p$ are high. We observe that this can sometimes lead to degenerate output, where the model generates an excessive number of points. 

To solve this, we add a special \textit{no-more-points} class with a fixed key embedding that the \patchtoken can attend to, meaning we have:
$$K_p = [W_{pk}\text{Norm}(h_i); h_{done}]$$
Where $h_{done}$ is a learned $M\times1$ vector. We use a position of 0 for $h_{done}$ when applying rotary embeddings.
If the no-more-points class is selected, the model is prevented from generating a \subpatchtoken and stops pointing.

\section{Training and Inference}
We train three models using this proposed method. We present high-level details of how they are trained but leave the specifics to the appendix.

\subsection{Implementation}
During pre-processing, we map input points to the corresponding target \imagetoken index, ViT patch index, and location index, and use those triples as additional input to the model. Our text input for points follows the Molmo2~\cite{molmo2} format, but replaces the string coordinates with the grounding tokens, including an additional \patchtoken at the end of each list of points that is assigned the no-more-points class. This reduces the number of tokens per coordinate from 8 (6 digits and 2 spaces) to 3.
For video, we also remove the text timestamps used by Molmo2 since they can be recovered based on which \imagetoken was selected, further reducing the token count.

As with Molmo2, we also give an integer object ID for each point, but place it after the coordinates instead of before.
Following Molmo2, we use a separate learning rate and gradient norm for the new pointing parameters. In general, we set the learning rate to match that used for the image-text connector parameters. We set $M=512$ for all experiments. In all training runs, we use packing and message-trees to support training on multiple examples per sequence~\cite{molmo2}.

\subsection{Inference}
During inference, we cache the keys of the image tokens and ViT patches during prefilling. This adds additional memory overhead, but the low-dimensionality of the keys means this uses roughly the same memory as the cached keys and values for 1-2 LLM layers, and it is only required for the image tokens.

We constrain the model to generate a \subpatchtoken and \locationtoken after each \patchtoken, and to only select \imagetokens that are the same, or after, any \imagetoken it has already selected in the input sequence, so output points are ordered correctly. We also prevent the model from generating multiple points with the same \imagetoken and ViT subpatch since we observe that this is almost always a case of the model pointing to the same thing twice. If the model selects the no-more-points class, we constrain the model to generate the \texttt{">} token, which ends a list of points in the Molmo2 pointing format.

To convert the selected patches back into coordinates, we retain a map of token\_id $\rightarrow$ coordinates for every ViT patch during pre-processing and combine it with the location predictions to get the output point.

\begin{figure}[t]
    \centering
    \includegraphics[width=0.99\linewidth]{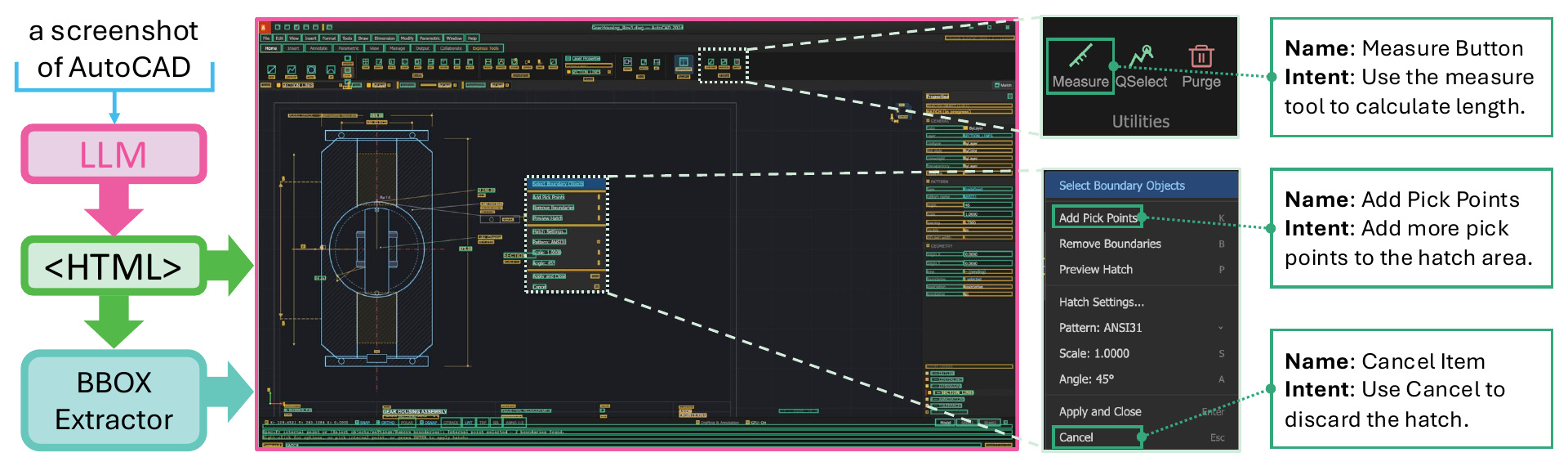}
    \caption{\textbf{Overview of the generation of \guidata{}.} We prompt an LLM to generate the HTML for the screenshot and extract all bounding boxes of its UI elements. Then we use LLMs to annotate each bounding box with its interaction intents.}
    \label{fig:molmop_syn}
\end{figure}

\subsection{Models}
\paragraph{\model{}.} We conduct a full end-to-end training run following the pipeline of Molmo2-8B. We use a larger batch size of 160 to better utilize the hardware we have available and lower the number of training steps from 30000 to 22,000 to compensate. To improve tracking, we also incorporate \trackdata{}, a new dataset of human-annotated and synthetic tracks (see below). We also slightly adjust the training mixture to better exploit the improved learning efficiency of the pointing data (see the appendix for details).

\paragraph{\modelImg{}.} The image pointing data in the Molmo2 mixture does not contain many instructional/GUI examples. To train a model better optimized for this task, we build \guidata, a code-guided synthetic GUI instructional dataset (see below for details), and fine-tune on it for 2000 steps with a batch size of 128 while increasing the image resolution to 48 crops per image.

\paragraph{\modelVid{}.} As with Molmo2, we observe that \model{} underperforms the specialized models on video grounding. We therefore also train a specialized video grounding model by finetuning \model{} after the pre-training stage on just video-pointing data for 6000 steps with a batch size of 64 and a max of 128 frames. We then fine-tune it for another 800 steps with a max of 384 frames to support longer videos.

\subsection{\guidata{}}
As shown in Figure \ref{fig:molmop_syn}, we extend the code-guided synthetic data generation framework (CoSyn) \cite{cosyn} to screenshot generation, in which we prompt the language model to generate HTML code that mimics digital environments for web, desktop, and mobile.
Given access to the underlying HTML code in each screenshot, we use the Playwright library with custom JavaScript to automatically extract bounding boxes for all elements in the screenshot.
We then feed the bounding box information to the language model to generate 5 pointing instructions per element that a user may ask when interacting with it.
In total, we synthesize 36K screenshots, with 2M densely annotated points and over 10M pointing instructions.
Qualitative examples of this data are provided in Figure \ref{fig:gui_example} in the appendix.

\begin{figure}[t]
    \centering
    \includegraphics[width=0.99\linewidth]{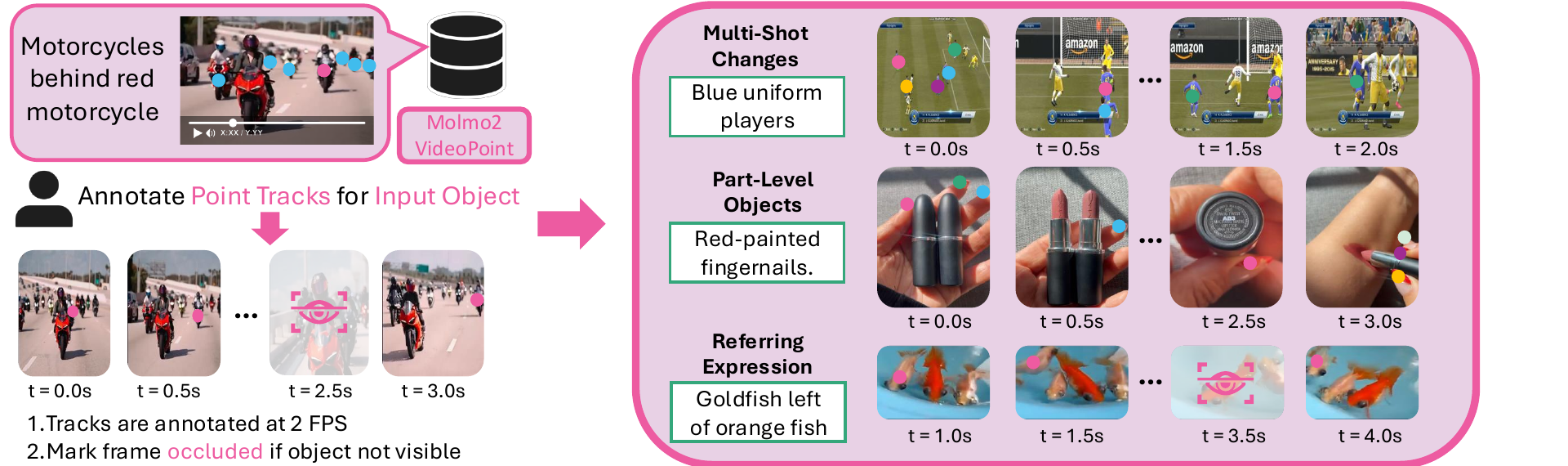}
    \caption{\textbf{\trackreal{}: human-annotated point-to-track extension.} Annotators are given a text query and an object of interest, and provide point tracks while marking frames as occluded when the object is not visible.}
    \label{fig:molmotrack_real}
\end{figure}

\subsection{\trackdata{}}
Existing tracking datasets with referring expressions, such as Molmo2-VideoTrack~\cite{molmo2}, were collected by expanding tracks for a fixed set of objects, resulting in limited scene and object diversity. Here, we contribute \trackdata{}, consisting of (1) \textbf{\trackreal{}}, human-annotated tracks on videos with any objects and (2) \textbf{\tracksynth{}}, synthetic tracks with diverse motion and occlusion patterns. For \trackreal{}, we extend Molmo2-VideoPoint annotations into full tracks via human annotation (Figure~\ref{fig:molmotrack_real}). For \tracksynth{}, we generate multi-object tracking videos in Blender with complex occlusion and motion dynamics, paired with automatically generated referring queries (Figure~\ref{fig:molmotrack_syn}). See Appendix~\ref{sect:trackdata} for collection details and qualitative examples.

\section{Results}

\subsection{Image Pointing}

\begin{table*}[!t]
\caption{\textbf{Point-Bench results.} Baseline scores taken from the Point-Bench leaderboard. Qwen3-VL-235B-A22B-Instruct and VisionReasoner-7B scores were taken from Poivre~\cite{poivre}, which did not include sub-category scores.}
\label{tab:point_bench}

\small
\centering
\begin{tabular}{@{}lcccccc@{}}
\toprule
\textbf{Model} & \textbf{Aff.} & \textbf{Spat.} & \textbf{Reason} & \textbf{Steer.} & \textbf{Count.} & \textbf{Avg} \\
\midrule
\color{gray}Human & \color{gray}92.3 & \color{gray}83.6 & \color{gray}87.8 & \color{gray}86.3 & \color{gray}95.6 & \color{gray}89.1 \\
\midrule
\apionlyheader{7} \\
Gemini-Robotics-ER-1.5~\cite{abdolmaleki2025gemini} & 69.7 & 69.7 & 60.1 & \textbf{67.5} & 68.5 & 67.1 \\
Gemini-2.5-Pro~\cite{comanici2025gemini} & 72.7 & 70.3 & 71.0 & 41.0 & 59.2 & 62.8 \\
\midrule
\openweightsheader{7} \\
Poivre-7B~\cite{poivre} & - & - & - & - & - & 67.5 \\
Qwen2.5-VL-32B-Instuct~\cite{qwen2} & 76.8 & 60.0 & 54.4 & 46.5 & 57.1 & 59.0 \\
Qwen2.5-VL-72B-Instuct~\cite{qwen2} & 76.8 & 60.0 & 54.4 & 46.5 & 57.1 & 59.0 \\
Qwen3VL~\cite{qwen3technicalreport} & 81.3 & 65.6 & 60.6 & 23.5 & 61.2 & 58.5 \\
Qwen3-VL-235B~\cite{qwen3technicalreport} & - & - & - & - & - &  58.3 \\

\midrule
\fullyopenheader{7}  \\
VisionReasoner-7B~\cite{liu2025visionreasoner} & - & - & - & - & - & 64.7 \\

{Molmo-7B-D~\cite{molmov1}} & 82.8 & 67.7 & 70.5 & 28.5 & 58.7 & 61.6 \\
{Molmo-72B~\cite{molmov1}} & 87.9 & 70.3 & 69.4 & 37.0 & 54.6 & 63.8 \\
{Molmo-7B-O~\cite{molmov1}} & 84.9 & 63.1 & 63.2 & 45.5 & 59.7 & 63.3 \\
{Molmo2-4B}\cite{molmo2} & 82.3 & 71.8 & 72.0 & 41.0 & 71.4 & 67.7 \\
{Molmo2-8B}\cite{molmo2} & 84.8 & 71.3 & 71.5 & 44.5 & 71.4 & 68.7 \\
{Molmo2-O-7B}\cite{molmo2} & 81.8 & 69.7 & 69.4 & 39.0 & 72.4 & 66.5 \\

\midrule
\modelheader{7} \\
\textcolor{molmocolor}{\model} & \textbf{85.9} & \textbf{76.9} & \textbf{77.2} & 39.0 & \textbf{74.5} & \textbf{70.7} \\
\bottomrule
\end{tabular}
\end{table*}

We show results on natural image pointing in Table~\ref{tab:point_bench} and Table~\ref{tab:pixmo_points}. \model{} is state-of-the-art on PointBench~\cite{pointarena}, surpassing Molmo2 by almost 2 points, including a 5 point gain in reasoning and spatial reasoning. On PixMo-Points~\cite{molmov1} \model{} surpasses Molmo2 by 4 points. Molmo2 and \model{} used the same data and training procedure, so these results show that using grounding tokens significantly boosts pointing capabilities on natural images.


\newcommand{\newcell}[1]{%
  \rotatebox{90}{%
    \parbox{1.7cm}{%
      \setlength{\baselineskip}{0.5em}%
      \textbf{\scriptsize{#1}}%
    }%
  }%
}

\begin{table*}[t]
\centering
\small
\setlength{\tabcolsep}{3.5pt}
\caption{\textbf{PixMo-Points results.} \model{} surpasses even proprietary models. We collect results for GPT-5.2, Gemini-3, and Qwen3-VL ourselves.}
\label{tab:pixmo_points}
\begin{tabular}{lccccccccccc}
\toprule
\textbf{Metric}
& \multicolumn{2}{c}{\textbf{API-only}}
& \multicolumn{2}{c}{\textbf{Open-weights}}
& \multicolumn{6}{c}{\textbf{Fully-open}}
& \multicolumn{1}{c}{\textbf{Ours}} \\
\cmidrule(lr){2-3} \cmidrule(lr){4-5} \cmidrule(lr){6-11} \cmidrule(lr){12-12}
& \newcell{GPT-5.2}
& \newcell{Gemini3-Pro}
& \newcell{Qwen3-VL-8B}
& \newcell{Qwen3-VL-4B}
& \newcell{Molmo-7B-D}
& \newcell{Molmo-72B}
& \newcell{Molmo-7B-O}
& \newcell{Molmo2-4B}
& \newcell{Molmo2-8B}
& \newcell{Molmo2-O-7B}
& \newcell{\textcolor{molmocolor}{\model}} \\
\midrule
Recall  & 31.0 & 77.3 & 54.3 & 45.1 & 76.4 & 74.9 & 74.4 & 83.3 & 85.5 & 83.1 & \textbf{90.4} \\
Precision  & 32.9 & 81.3 & 53.5 & 44.2 & 76.2 & 74.9 & 74.6 & 85.1 & 86.4 & 83.7 & \textbf{89.3} \\
F1 & 31.6 & 77.8 & 53.4 & 43.7 & 75.7 & 74.5 & 74.0 & 83.4 & 85.2 & 82.7 & \textbf{89.2} \\
\bottomrule
\end{tabular}
\end{table*}

\subsection{GUI Pointing}
\begin{table*}[!t]
\caption{\textbf{GUI grounding results.} \modelImg is our GUI specialized model finetuned on \guidata{} (Synthetic GUI dataset) we constructed. \textit{Molmo2-GUI-8B} also fine-tunes on \guidata{} but without our token referencing mechanism. (64crops) denotes our test-time scaling for inference with more image crops. The best performance of \textit{fully open} models is \textbf{bold}. Scores with * are from evaluations in~\cite{venusteam2026uivenus15technicalreport,qin2025ui}.}
\label{tab:computer_user}

\centering
\small
\begin{tabular}{lcccc}
\toprule
\textbf{Model} & \textbf{ScreenSpot-V2} & \textbf{ScreenSpot-Pro} & \textbf{OSWorldG} \\
\midrule
\apionlyheader{4}  \\ 
Claude 3.7 \cite{anthropic2024claude} & 87.6 & 27.7 & - \\
OpenAI CUA \cite{openai2024cua} & 87.9 & 23.4 & - \\ 
Gemini-3-Pro \cite{gemini3} & 93.7 & 72.7  & 35.5 \\ \midrule
\openweightsheader{4}  \\ 
Holo2-8B \cite{hai2025holo2modelfamily} & 93.2 & 58.9 & 70.1 \\ 

UI-TARS 1.5-7B \cite{qin2025ui}& 94.2 & 61.6 & 64.2\rlap{$^*$} \\

UI-Venus-1.5-8B \cite{venusteam2026uivenus15technicalreport} & 95.9 & 68.4 & 69.7 \\

Qwen3-VL-8B \cite{bai2025qwen3vltechnicalreport} & 92.1\rlap{$^*$} & 52.7\rlap{$^*$} & 57.5\rlap{$^*$} \\

MAI-UI-8B \cite{zhou2025maiuitechnicalreportrealworld} & 95.2 & 65.8 & 60.1 \\


\midrule
\fullyopenheader{4}  \\ 
GUI-Actor \cite{wu2025gui} & 90.9 & 41.8 & - \\
JEDI-7B \cite{jedi} & 91.7& 50.2 &  54.1 \\
GroundCUA-7B \cite{groundcua}  & 89.3 & 50.2 & 67.2 \\
OpenCUA-7B \cite{wang2025opencua} & 92.3 & 50.0 & 55.3 \\
GTA1-7B \cite{GTA1-7B} & 92.4 & 50.1 & 60.1 \\
Molmo2-8B\cite{molmo2} & 89.5 & 30.4 & 54.1 \\
\textcolor{molmocolor}{\textit{Molmo2-GUI-8B}} & 88.8 & 52.3 & 66.1 \\
\midrule    
\modelheader{4} \\ 
\textcolor{molmocolor}{\model} & 89.8 & 36.4 & 54.9 \\
\textcolor{molmocolor}{\model (64crops)} & 89.8 & 39.4 & 56.5 \\
\textcolor{molmocolor}{\modelImg} & 93.4 & 60.2 & \textbf{70.0} \\
\textcolor{molmocolor}{\modelImg (64crops)} & \textbf{93.9} & \textbf{61.1} & \textbf{70.0} \\
\bottomrule
\end{tabular}
\end{table*}
We show results on ScreenSpot-V2~\cite{li2025screenspot}, ScreenSpot-Pro~\cite{li2025screenspot}, and OSWorldG~\cite{xie2025scalingcomputerusegroundinguser}. In addition to other models, we also compare to a baseline, Molmo2-GUI-8B, built by fine-tuning Molmo2-8B on the same data \modelImg was trained on.
The Molmo2 data mixture does not contain instruction-point pairs, so \model sometimes does not point when given them as input. To fix this, we use constrained decoding for both the Molmo2 models and \model{} (but not \modelImg{}) to force the model to generate exactly one point. We also show results with \textit{test-time-scaling} where we increase the number of crops during test time to 64. We find that test-time scaling breaks models that use text coordinates, dropping performance to $<10\%$, presumably due to the model not knowing how to map the larger number of patches to text coordinates. Therefore, we do not use it for other models.

Results are shown in Table~\ref{tab:computer_user}.
Compared to Molmo2, \model{} is even on ScreenSpot-V2 but shows significant improvements on ScreenSpotPro and OSWorldG, again showing the benefit of our pointing method.
Fine-tuning with instruction-image data makes \modelImg SoTA among fully open models on all tasks. Open-weight models UI-Venus and MAI-UI show better results, likely because both models utilize large-scale proprietary data collection efforts as well as more elaborate training pipelines that include RL. 

We observe a gap of 2 to 9 points between \modelImg and the baseline that uses text coordinates on ScreenSpotPro, showing that our model design is critical for this high performance. We hypothesize that the large gap of 9 points in ScreenSpotPro might be due to grounding tokens having a particularly high impact when dealing with high-resolution input.

\subsection{Video Pointing}
\newcommand{\mycellg}[2]{%
  \rotatebox{90}{%
    \parbox{2.8cm}{%
      \setlength{\baselineskip}{0.5em}%
      \textbf{\scriptsize{#1}}\\
      \footnotesize{\textcolor{gray}{#2}}%
    }%
  }%
}

\begin{table}[!t]
    \caption{\textbf{Video counting and pointing results.} 
    \model\ scores highest on BURST-VC and \model-VP and second highest on \model-VC's close accuracy, slightly behind Gemini 2.5 Pro. Best open model results are \textbf{bold}.}
    \label{tab:video_count_and_point_results}
    
    \centering
    \small
    \begin{tabular}{@{}lcccccccc@{}}
    \toprule
     & 
     
        \multicolumn{2}{@{}c}{\textbf{BURST VC (test)~\cite{athar2023burst}}} 
        & \multicolumn{3}{@{}c}{\textbf{Molmo2-VC}} 
        & \multicolumn{3}{@{}c}{\textbf{Molmo2-VP}} \\ 
       \textbf{Model}  & 
       Acc.
        & Close acc.
        & Acc.
        & Close acc.
        & F1 & Recall & Precision\\
        \midrule
        \apionlyheader{7} \\
        GPT-5~\cite{gpt5}             & 43.1	& 73.7 & 35.8 & 50.3 & 4.1 &  4.4 & 4.2  \\
        GPT-5 mini~\cite{gpt5}         & 46.0 & 73.0 & 29.8 & 49.3 & 2.2 & 2.2 & 2.2  \\
          Gemini 3 Pro~\cite{gemini3}    & 44.0 & 71.7 & 37.1 & 53.1 & 20.0 & 27.4 & 19.8\\
        Gemini 2.5 Pro~\cite{comanici2025gemini}    & 41.6 & 70.0 & 35.8 & 56.5 & 13.0 & 14.5 & 13.6\\
        Gemini 2.5 Flash~\cite{comanici2025gemini}   & 38.7 & 70.0 & 31.9 & 48.2 & 11.1 & 11.2 & 12.2 \\
        Claude Sonnet 4.5~\cite{anthropic2025sonnet}  & 42.4 & 72.6 & 27.2 & 45.1 & 3.5 & 3.7 & 4.3  \\
        \midrule
        \openweightsheader{7} \\
        Qwen3-VL-4B~\cite{bai2025qwen3vltechnicalreport}        & 38.9 & 74.7 & 25.3 & 44.3 & 0.0 & 0.0 & 0.0 \\
        Qwen3-VL-8B~\cite{bai2025qwen3vltechnicalreport}         & 42.0 & 74.4 & 29.6 & 47.7 & 1.5 & 1.5 & 1.5\\
        \midrule
        \fullyopenheader{7} \\
        {Molmo2-4B} & 61.5 & 76.1 & 34.3 & 56.1 & \textbf{39.9} & \textbf{42.7} & \textbf{39.4} \\
        {Molmo2-8B}    & 60.8 & 75.0 & 35.5 & 53.3 & {38.4} & {39.3} & 38.7 \\
        {Molmo2-O-7B}    & 61.6 & 76.0 & 33.2	& 50.5 & 35.8 & 35.8	 & 37.9 \\
        \midrule
        \modelheader{7} \\
        \textcolor{molmocolor}{\model} & 61.6 & \textbf{76.9} & 35.6 & 54.6 & 36.2 & 35.7 & 37.8 \\
        \textcolor{molmocolor}{\modelVid} & \textbf{62.0} & 76.3 & \textbf{36.0} & \textbf{58.7} & 38.8 & 39.8 & 38.8
        \\ \bottomrule
    \end{tabular}
\end{table}

We evaluate video pointing on BURST-VideoCount~\cite{athar2023burst},  Molmo2-VideoCount, and Molmo2-VideoPoint~\cite{molmo2}. For the counting datasets, we report accuracy as well as close accuracy, which measures if the number of points is almost correct (computed as $|pred - gt| \leq \Delta$, where $\Delta = 1 + \lfloor 0.05\times gt \rfloor$). 
For Molmo2-VideoPoint, we report F1, recall, and precision metrics when matching points to ground-truth segmentation masks. Baseline numbers come from Molmo2~\cite{molmo2}.

For \model{}, we see an improvement on both Burst-VC and Molmo2-VC compared to Molmo2-8B, although we also see a drop in Molmo2-VP. To get a more definitive result, we conduct a human preference evaluation using predictions from both models on 470 video pointing queries (See appendix for details). We find 152 ties, 130 wins for Molmo2, and 188 wins for \model{}. Excluding ties, \model{} has a 59.1\% win rate, showing humans prefer \model{}'s output.
\modelVid{} sees a more consistent gain, including a full 5 point gain on Molmo2-VC close, surpassing Gemini 3 Pro.

\subsection{Tracking Results}
\label{sect:track}
\begin{table*}[!t]
    \caption{\textbf{Tracking Results on Academic Benchmark}. $\mathcal{J}\&\mathcal{F}$ measures the segmentation mask quality of object tracks. F1 measures whether points fall in the mask, and HOTA~\cite{luiten2021hota} further accounts for consistent ID associations. }

    \renewcommand{\arraystretch}{0.9}
    \centering
    \setlength{\tabcolsep}{2.5pt}
    \resizebox{\textwidth}{!}{
    \begin{tabular}{@{}lccccccccccccccccc@{}}
    \toprule
        &{\textbf{MeViS~\cite{ding2023mevis}}} & \multicolumn{3}{@{}c}{\textbf{MeViS~\cite{ding2023mevis}}} & \multicolumn{3}{@{}c}{\textbf{Ref-YT-VOS~\cite{seo2020urvos}}} & \multicolumn{3}{@{}c}{\textbf{Ref-Davis~\cite{refdavis}}} & \multicolumn{3}{@{}c}{\textbf{ReasonVOS~\cite{bai2024one}}} \\
        & valid & \multicolumn{3}{@{}c}{valid-u} & \multicolumn{3}{@{}c}{valid} & \multicolumn{3}{@{}c}{valid} &
        \multicolumn{3}{@{}c}{test} \\
        \textbf{Model} & $\mathcal{J}\&\mathcal{F}$ & $\mathcal{J}\&\mathcal{F}$ & F1 & HOTA &  $\mathcal{J}\&\mathcal{F}$ & F1 & HOTA &  $\mathcal{J}\&\mathcal{F}$ & F1 & HOTA &  $\mathcal{J}\&\mathcal{F}$ & F1 & HOTA \\ 
        \midrule
        
        \multicolumn{10}{@{}l}{\textbf{\textit{API call only}}} \\
        GPT-5~\cite{gpt5}                            & 23.4 & 26.5 & 17.3 & 14.0 & 30.9 & 21.0 & 18.4 & 25.2 & 17.0 & 11.6 & 24.7 & 13.6 & 10.7 \\
        GPT-5 mini~\cite{gpt5}                       & 15.7 & 15.4 & 8.5  & 6.8  & 16.2 &  7.4 & 6.2  & 8.4  & 3.4  & 2.3  & 14.6 & 4.2  & 3.4 \\
        Gemini 3 Pro~\cite{gemini3}                  & 42.5 & 51.1 & 42.3 & 36.0 & 55.0 & 49.1 & 45.5 & 66.6 & 60.8 & 55.7 & 52.6 & 48.5 & 42.1\\ 
        Gemini 2.5 Pro~\cite{comanici2025gemini}     & 40.7 & 52.8 & 41.2 & 35.0 & 45.1 & 44.5 & 40.5 & 45.6 & 62.7 & 56.6 & 44.0 & 50.2 & 42.4 \\
        Gemini 2.5 Flash~\cite{comanici2025gemini}   & 27.6 & 31.8 & 24.0 & 19.9 & 36.0 & 32.8 & 30.0 & 31.6 & 36.7 & 30.0 & 26.5 & 25.8 & 21.0 \\     
        
        \midrule

        \multicolumn{10}{@{}l}{\textbf{\textit{Open weights only}}} \\
        Qwen3-VL-4B~\cite{qwen3technicalreport} & 29.7 & 30.6 & 23.3 & 18.7 & 32.1 & 29.0 & 26.5 & 44.4 & 33.1 & 26.9 & 26.5 & 17.0 & 13.5 \\
        Qwen3-VL-8B~\cite{qwen3technicalreport} & 35.1 & 34.4 & 30.1 & 23.8 & 48.3 & 42.1 & 37.6 & 41.0 & 41.6 & 33.2 & 24.9 & 22.3 & 17.5 \\
        \midrule
        
        \multicolumn{10}{@{}l}{\textbf{\textit{Specialized open models}}} \\
        VideoLISA~\cite{bai2024one}                      & 44.4 & 53.2 & -- & -- & 63.7 & -- & -- & 68.8 & -- & -- & 47.5  & -- & -- \\
        VideoGLaMM~\cite{rasheed2024glamm}               & 45.2 & 50.6 & -- & -- & 66.8 & -- & -- & 69.5 & -- & -- & 33.9 & -- & -- \\
        Sa2VA-8B~\cite{yuan2025sa2va}                    & 46.9 & 57.0 & -- & --  & \textbf{70.7} & -- & --  & \underline{75.2} & -- & -- & 55.5 & -- & -- \\
        Sa2VA-Qwen3-VL-4B~\cite{yuan2025sa2va}           & 36.7 & 57.1 & -- & --  & 68.1 & -- & -- &  \textbf{76.0}  & -- & -- & 50.0 & -- & -- \\
        
    \midrule
    \multicolumn{10}{@{}l}{\textbf{\textit{Fully open}}} \\
    Molmo~\cite{molmov1} + SAM 2~\cite{ravi2024sam2} & 46.9 & 51.5 & 53.8 & -- & 64.6 & 71.1 & -- & 65.2 & 74.5 & -- & 45.7 & 50.3 & -- \\
    VideoMolmo-7B~\cite{ahmad2025videomolmo}            & 53.9 & 57.0 & 59.4 & -- & 67.3 & 73.7 & -- & 72.5 & 75.4 & -- & 51.1 & 50.3 & -- \\
    \textcolor{molmocolor}{Molmo2-4B}          &  \underline{63.3} & 70.0 & 75.5 & 72.4 & 70.2 & \underline{80.4} & \underline{78.8} & 73.5 & \underline{83.1} & \underline{81.1} & 61.9 & 66.5 & 64.0 \\ 
    \textcolor{molmocolor}{Molmo2-8B}          &  62.3 & \underline{70.8} & 75.9 & \underline{72.6} & 70.2 & \underline{78.7} & \underline{77.3} & 72.7 & 81.3 & 78.7 & \textbf{65.8} & \textbf{70.8} & \textbf{68.6} \\ 
    \textcolor{molmocolor}{Molmo2-O-7B}        &  58.4  & 69.7 & \underline{76.1} & 72.3 & 67.9 & 77.7 & 76.1 & 70.4 & 79.2 & 76.0 & 62.6 & 67.5 & 65.1 \\ 

    \midrule
    \modelheader{13} \\
    \textcolor{molmocolor}{\model{}} & \textbf{63.5} & \textbf{72.2} & \textbf{77.0} & \textbf{73.8} & \underline{70.5} & \textbf{81.9} & \textbf{80.5} & 73.6 & \textbf{84.1} & \textbf{82.2} & \underline{64.7} & \underline{68.8} & \underline{67.0} \\ \bottomrule
        
    \end{tabular}
    }
    
    \label{tab:track_result_academic}
\end{table*}

\begin{table*}[!t]
    \caption{\textbf{Results on Molmo2Track benchmark by video domain}. Overall reports the accuracy across all samples.}    

    \renewcommand{\arraystretch}{0.98}
    \centering
    \setlength{\tabcolsep}{2.5pt}
    \resizebox{\textwidth}{!}{
    \begin{tabular}{@{}l ccc ccc ccc ccc ccc| ccc@{}}
    \toprule
        & \multicolumn{3}{@{}c}{\textbf{Animals}}   & \multicolumn{3}{@{}c}{\textbf{Person}} & \multicolumn{3}{@{}c}{\textbf{Sports}} & \multicolumn{3}{@{}c}{\textbf{Dancers}} & \multicolumn{3}{@{}c}{\textbf{Misc}} & \multicolumn{3}{@{}c}{\textbf{Overall}}\\
        \textbf{Model} & $\mathcal{J}\&\mathcal{F}$ & F1 & HOTA &  $\mathcal{J}\&\mathcal{F}$ & F1 & HOTA &  $\mathcal{J}\&\mathcal{F}$ & F1 & HOTA &  $\mathcal{J}\&\mathcal{F}$ & F1 & HOTA &  $\mathcal{J}\&\mathcal{F}$ & F1 & HOTA &  $\mathcal{J}\&\mathcal{F}$ & F1 & HOTA \\
        \midrule
        
        \multicolumn{10}{@{}l}{\textbf{\textit{API call only}}} \\
        GPT-5~\cite{gpt5}                            & 41.4 & 20.6 & 20.3 & 16.5 & 4.5  & 4.2 & 14.4 & 2.0 & 2.5 & 33.8 & 11.7 & 11.5 & 14.6 & 2.2 & 1.6 & 23.5 & 7.5 & 7.5 \\
        GPT-5 mini~\cite{gpt5}                       & 21.7 & 7.8  & 8.0  & 8.6  & 1.6  & 1.5 & 10.7 & 0.6 & 0.8 & 15.6 & 2.1  & 2.0  & 13.5 & 0.6 & 0.4 & 12.7 & 2.1 & 2.1 \\
        Gemini 3 Pro~\cite{comanici2025gemini}       & 70.4 & 62.3 & 60.0 & 44.5 & 30.7 &29.2 & 23.4 & 10.3& 8.8 & 55.6 & 44.3 & 37.8 & 35.3 & 18.3&	14.4 & 44.6 & 32.2 & 29.1\\
        Gemini 2.5 Pro~\cite{comanici2025gemini}     & 69.3	& 56.8 & 53.2 & 50.0 & 33.6	& 31.9 & 29.7 & 10.8 & 8.9 & 55.9& 39.4 & 32.2 & 34.7 & 17.6 & 18.3 & 47.9 & 31.2 & 27.8\\
        Gemini 2.5 Flash~\cite{comanici2025gemini}   & 58.0 & 46.6 & 44.4 & 38.9 & 21.4	& 20.1 & 13.2 & 6.2	 & 5.5 & 48.0 & 29.0 & 25.1	& 21.9 & 5.7 & 4.6 & 36.2 & 21.8 & 19.8 \\        
        \midrule

        \multicolumn{10}{@{}l}{\textbf{\textit{Open models}}} \\
        Qwen3-VL-4B~\cite{qwen3technicalreport} & 57.2 & 11.5 & 12.3 & 35.1 & 12.0 & 11.2 & 3.8 & 0.4 & 0.4 & 34.6 & 6.9 & 5.7 & 17.5 & 6.2 & 4.2 & 28.5 & 7.2 & 6.7\\
        Qwen3-VL-8B~\cite{qwen3technicalreport} & 63.8 & 52.3 & 50.2 & 35.4 & 20.3 & 18.9 & 5.2 & 1.7 & 1.4 & 31.3 & 19.0 & 16.7 & 16.3 & 6.2 & 4.2 & 28.7 & 18.0 & 16.5\\
        \midrule
        
        \multicolumn{10}{@{}l}{\textbf{\textit{Specialized open models}}} \\
     
        VideoLISA~\cite{bai2024one}              & 67.8 & -- & -- & 35.8 & -- & -- & 32.9 & -- & --  & 53.6 & -- & -- & 25.8 & -- & -- & 43.3 & -- & --\\
        VideoGLaMM~\cite{rasheed2024glamm}       & 63.9 & -- & -- & 26.2 & -- & -- & 34.3 & -- & -- & 46.0 & -- & -- & 22.3 & -- & -- & 37.9 & -- & --\\
        Sa2VA-8B~\cite{yuan2025sa2va}            & 74.3 & -- & -- & 45.5 & -- & -- & 30.7 & -- & -- & 53.3 & -- & -- & \textbf{49.1} & -- & -- & 46.9 & -- & -- \\
        Sa2VA-Qwen3-VL-4B~\cite{yuan2025sa2va}   & 73.3 & -- & -- & 48.6 & -- & -- & 31.6 & -- & -- & 50.1 & -- & -- & 31.4 & -- & -- & 46.7 & -- & -- \\
        SAM 3~\cite{carion2025sam} & 41.1 & -- & -- & 35.2 & -- & -- & 43.3 & -- & -- & 29.2 & -- & -- & 36.8 & -- & -- & 36.3 & -- & -- \\
        Molmo~\cite{molmov1} + SAM 2~\cite{ravi2024sam2} & 71.8 & 76.0 &  -- & \underline{52.7} & 7.0 & -- & 52.8 & 2.6 & -- & 51.7 & 7.55 & -- & 40.9 & 37.5 & -- & 54.2 & 14.0 & -- \\
        VideoMolmo-7B~\cite{ahmad2025videomolmo} & 68.4 & 69.5 & -- & 51.1 & 6.3 & -- & 43.2 & 2.1 & -- & 53.8 & 7.2 & -- & 39.9 & 30.8 & -- & 51.3 & 12.7 & -- \\
        
        \midrule
        \multicolumn{10}{@{}l}{\textbf{\textit{Fully Open}}} \\
        \textcolor{molmocolor}{Molmo2-4B}          &  \underline{81.0} & \underline{83.0} & \underline{83.7} & 43.7 & \underline{48.3} & 47.7 & 59.7 & 53.1 & 54.3 & \underline{60.4} &  \underline{64.4} & \textbf{64.4} & 43.1 & 35.1 & 31.3 & \underline{56.7} & \underline{57.5} & \underline{57.6} \\ 
        \textcolor{molmocolor}{Molmo2-8B}          &  80.1 & 82.0 & 83.0 & 43.1 & 47.9 & \underline{48.0} & \underline{59.8} & \underline{53.3} & \underline{54.8} & 59.9 & 63.9 & 63.5 & 41.6 & 31.5 & 29.7 & 56.2 & 57.1 & 57.5 \\ 
        \textcolor{molmocolor}{Molmo2-O-7B}        &  80.1 & 81.9 & 82.8 & 41.5 & 45.5 & 45.4 & 54.1 & 47.6 & 48.6 & 57.7 & 61.0 & 60.3 & \underline{45.0} & \underline{37.6} & \underline{34.7} & 53.7 & 54.2 & 54.2 \\

    \midrule
    \modelheader{18} \\
    \textcolor{molmocolor}{\model{}} & \textbf{81.9} & \textbf{84.4} & \textbf{85.8} & \textbf{56.9} & \textbf{51.2} & \textbf{50.9} & \textbf{63.6} & \textbf{56.9} & \textbf{57.7} & \textbf{61.8} & \textbf{65.6} & \underline{64.2} & \textbf{49.1} & \textbf{45.7} & \textbf{43.1} & \textbf{62.5} & \textbf{60.2} & \textbf{60.0} \\ \bottomrule
    \end{tabular}
    }
    \label{tab:molmo2_track_results}
\end{table*}
We evaluate \model{} on the tracking benchmarks introduced in Molmo2~\cite{molmo2}. Table~\ref{tab:track_result_academic} presents results on academic benchmarks and Table~\ref{tab:molmo2_track_results} reports results on \model-Track across video domains. Following~\cite{ahmad2025videomolmo, molmo2}, 
 Jaccard and F-measure ($\mathcal{J}\&\mathcal{F}$), which measures segmentation quality, is computed by passing \model{} points as input to SAM2~\cite{ravi2024sam2} to obtain segmentation masks. F1 and HOTA~\cite{luiten2021hota} evaluate point accuracy directly, where F1 measures whether predicted points fall within the ground-truth segmentation and HOTA further accounts for association consistency across frames.

Overall, \model{} shows substantial gains over Molmo2-8B across tracking benchmarks. Notable improvements come from the multi-object tracking dataset as \model{} reaches \textbf{63.5 vs. 62.3} and \textbf{72.2 vs. 70.8 $\boldsymbol{\mathcal{J}}\&\boldsymbol{\mathcal{F}}$} on MeViS valid and valid-u splits~\cite{ding2023mevis}. On Molmo2-Track, the gains are consistent across all video domains, with overall improvements of \textbf{+5.7 $\boldsymbol{\mathcal{J}}\&\boldsymbol{\mathcal{F}}$},  \textbf{+3.1 F1} and \textbf{+2.5 HOTA}. We suspect our grounding tokens enable better grounding and instance-level identification in tracking as well. One exception is ReasonVOS~\cite{bai2024one} which targets queries that require semantic reasoning rather than precise spatial grounding, thus grounding tokens provide fewer benefits here. 

\subsection*{Tracking Ablations}
To disentangle our modeling and data contributions, we conduct ablations that sequentially remove grounding tokens and \trackdata{}. Due to computation constraints, all ablation models are trained for 5K steps on heavily upsampled tracking data. As shown in Table~\ref{tab:track_ablations}, both contribute meaningfully to tracking quality. Removing grounding tokens leads to a  \textbf{drop of 4.6 F1} and \textbf{4.0 HOTA} overall, suggesting grounding tokens are particularly beneficial for tracking a diverse set of object types. 
Further removing \trackdata{} yields additional losses, most notably on Misc (\textbf{-7.6 F1}), confirming that the expanded data coverage addresses gaps where the original training data had limited representation.

\begin{table}[!t]
    \caption{\textbf{Tracking ablations on Molmo2Track}. We consecutively remove grounding tokens and the newly introduced \trackdata{} data.}
    \centering
    \small
    \resizebox{\columnwidth}{!}{
    \begin{tabular}{l cc cc cc cc cc cc}
    \toprule
    \textbf{Model} & \multicolumn{2}{c}{\textbf{Animals}} & \multicolumn{2}{c}{\textbf{Person}} & \multicolumn{2}{c}{\textbf{Sports}} & \multicolumn{2}{c}{\textbf{Dancers}} & \multicolumn{2}{c}{\textbf{Misc}} & \multicolumn{2}{c}{\textbf{Overall}} \\
     & F1 & HOTA & F1 & HOTA & F1 & HOTA & F1 & HOTA & F1 & HOTA & F1 & HOTA \\ \midrule
    \textcolor{molmocolor}{\model{}-Ablation} & 83.4 & 85.3 & 49.2 & 49.1 & 56.5 & 57.6 & 64.9 & 63.7 & 47.5 & 44.5 & 59.3 & 59.3 \\
    \midrule
    w/o grounding tokens & 80.7 & 81.8  & 44.4 & 44.5 & 49.4 & 51.4 & 62.9 & 62.9 & 37.9 & 35.5 & 54.7 & 55.3 \\
    w/o \trackdata{} & 80.0 & 81.0 & 43.5 & 43.8 & 49.4 & 50.5 & 61.8 & 61.6 & 30.3 & 28.1 & 53.8 & 54.2 \\
    \bottomrule
    \end{tabular}
    }
    \label{tab:track_ablations}
\end{table}

\subsection{MultiTask Results}

\begin{table*}[t]
\caption{\textbf{Image and video QA results}. Averaged results on the QA benchmarks in~\cite{molmo2}, see the appendix for details. Some results were computed by \cite{molmo2}.}
\label{tab:qa}

\centering
\small
\begin{tabular}{lcccc}
\toprule
\textbf{Model} & \makecell{\textbf{Image} \\ \textbf{Avg.}} & \makecell{\textbf{Multi-Image} \\ \textbf{Avg.}} & \makecell{\textbf{Short Video} \\ \textbf{Avg.}} & \makecell{\textbf{Long Video} \\ \textbf{Avg.}} \\ \midrule
\apionlyheader{5}  \\
GPT-5\cite{gpt5} & 83.7 & 72.1 & 73.1 & 76.3 \\
GPT-5 mini\cite{gpt5} & 81.9 & 68.2 & 66.8 & 69.8 \\
Gemini 3 Pro\cite{gemini3} & 86.2 & 81.9 & 71.0 & 78.8 \\
Gemini 2.5 Pro\cite{comanici2025gemini} & 81.3 & 72.4 & 71.1 & 80.4 \\
Gemini 2.5 Flash\cite{comanici2025gemini} & 79.3 & 68.3 & 67.0 & 74.5 \\
Claude Sonnet 4.5\cite{anthropic2025sonnet} & 76.3 & 59.5 & 62.8 & 66.4 \\
\midrule
\openweightsheader{5}  \\
InternVL3.5-4B\cite{wang2025internvl3} & 77.2 & 53.5 & 62.0 & 56.5 \\
InternVL3.5-8B\cite{wang2025internvl3} & 78.2 & 54.9 & 63.0 & 57.1 \\
Qwen3-VL-4B\cite{bai2025qwen3vltechnicalreport} & 78.4 & 57.6 & 63.7 & 62.7 \\
Qwen3-VL-8B\cite{bai2025qwen3vltechnicalreport} & 81.2 & 56.3 & 65.3 & 63.5 \\
Keye-VL-1.5-8B\cite{yang2025kwai} & 79.8 & 52.1 & 60.1 & 60.4 \\
GLM-4.1V-9B\cite{glmv} & 77.0 & \textbf{67.4} & 64.2 & 60.5 \\
MiniCPM-V-4.5-8B\cite{yu2025minicpmv45cookingefficient} & 77.7 & 47.3 & 62.1 & 60.1 \\
Eagle2.5-8B\cite{eagle2_5} & 81.2 & 52.0 & 67.0 & \textbf{65.2} \\
\midrule
\fullyopenheader{5}  \\
PLM-3B\cite{cho2025PerceptionLM} & 75.9 & 40.6 & 66.3 & 53.5 \\
PLM-8B\cite{cho2025PerceptionLM} & 78.7 & 35.7 & 68.5 & 56.2 \\
LLaVA-Video-7B\cite{llava_video} & - & - & 59.4 & 56.2 \\
VideoChat-Flash-7B\cite{li2024videochat} & - & - & 66.4 & 58.1 \\
Molmo2-4B\cite{molmo2} & 80.4 & 57.8 & 69.3 & 64.5 \\
Molmo2-8B\cite{molmo2} & 81.7 & 56.4 & \textbf{69.9} & 64.1 \\
Molmo2-O-7B\cite{molmo2} & 79.7 & 53.5 & 68.1 & 59.2 \\
\midrule
\modelheader{5}  \\
\textcolor{molmocolor}{\model} & \textbf{82.2} & 56.0 & 69.5 & 64.2  \\ \bottomrule
\end{tabular}
\end{table*}
To evaluate how well \model{} works as a general-purpose VLM, Table~\ref{tab:qa} reports average performance on 11 image benchmarks~\cite{ai2_diagram, chartqa, mathew2021docvqa, infoqa, textqa, goyal2017making, realworldqa, yue2024mmmu, lu2024mathvista, beyer2024paligemma, molmov1}, 3 multi-image benchmarks~\cite{wang2024muirbench, meng2024mmiumultimodalmultiimageunderstanding, fu2024blink}, 6 short-video benchmarks~\cite{nextqa, perception_test, mvbench, tomato, motionbench, tempcompass}, and 7 long video benchmarks~\cite{videomme, longvideobench, mlvu, lvbench, videoevalpro, egoschema} following the evaluation protocol Molmo2. Full details are in the appendix.


Overall, we see only small changes compared to Molmo2-8B: a small gain on images, a small loss on multi-images and short videos, and no difference on long videos. 
Since the data and training protocol are mostly unchanged between the two models, better transfer from image pointing data to image tasks might explain the gain in images. Although some interference between grounding and QA might likewise explain the drop in short videos. We do observe another sign of positive transfer during pre-training, where captioning performance is slightly better for \model{} after pre-training (55.6 vs 55.9 F1 on the Molmo captioning metric~\cite{molmov1}). However, in either case, the effect is small.

\begin{table*}[!t]
\caption{\textbf{Ablations}. Results when removing rotary embeddings, the no-more-points class, or the constraint that the points must be generated in sorted order.}

\centering
\small
\renewcommand\theadfont{\scriptsize\bfseries}
\begin{tabular}{lccccccc}
\toprule
\textbf{Model} & \textbf{PixMoPoint} & \textbf{PointBench} & \multicolumn{3}{c}{\textbf{Molmo2-VC}} \\
 & \textbf{F1} & \textbf{Avg.} & \textbf{Correct} & \textbf{Close} & \textbf{Overcount} $\downarrow$\\ \midrule
\textcolor{molmocolor}{Molmo2-P-Ablation} & 85.2 & 67.8 & 58.0 & 36.6 & 3.6 \\
\midrule
w/o rotary & 84.5 & 67.6 & 56.8 & 33.0 & 4.5 \\
w/o no-more-points & 84.7 & 66.6 & 52.3 & 32.8 & 10.3 \\
w/o point sorting & 83.6 & 71.2 & 40.0 & 24.4 & 3.2 \\ \bottomrule
\end{tabular}
\label{tab:ablations}
\end{table*}
\subsection{Modeling Ablations}
Modeling ablations are shown in Table~\ref{tab:ablations}. 
Due to limited compute resources, we do ablations with a lighter-weight training pipeline that starts from a pre-trained captioning model (without any pointing capabilities) and then fine-tunes on the image and video pointing data from the Molmo2 data mixture. We train for 6000 steps with a batch size of 64.
For video input, we observe that models can produce degenerate outputs with an excessive number of points. To capture this, we add an \textit{overcount} metric, defined as how often predictions are $>10$ points and $>=$ twice the number of ground truth points.

Removing rotary embeddings decreases performance a bit on images and has a more significant effect on video. 
Removing the no-more-points token hurts performance and more than doubles the amount of overcounting.
Randomizing the order of the points shows a significant drop for video, but a surprising gain on PointBench. We hypothesize that this might improve performance by allowing models to generate points in an ``easiest point first" order~\cite{chang2022maskgit}, but more work will be needed to take advantage of this without degrading video.

\subsection{Sample Efficiency}
\begin{figure}[!t]
\scriptsize
\centering
\caption{\textbf{Sample efficiency}. Left: Performance when using a very limited number of pointing training examples. Right: Pointing performance during full-scale pre-training.}
\includegraphics[width=\linewidth]{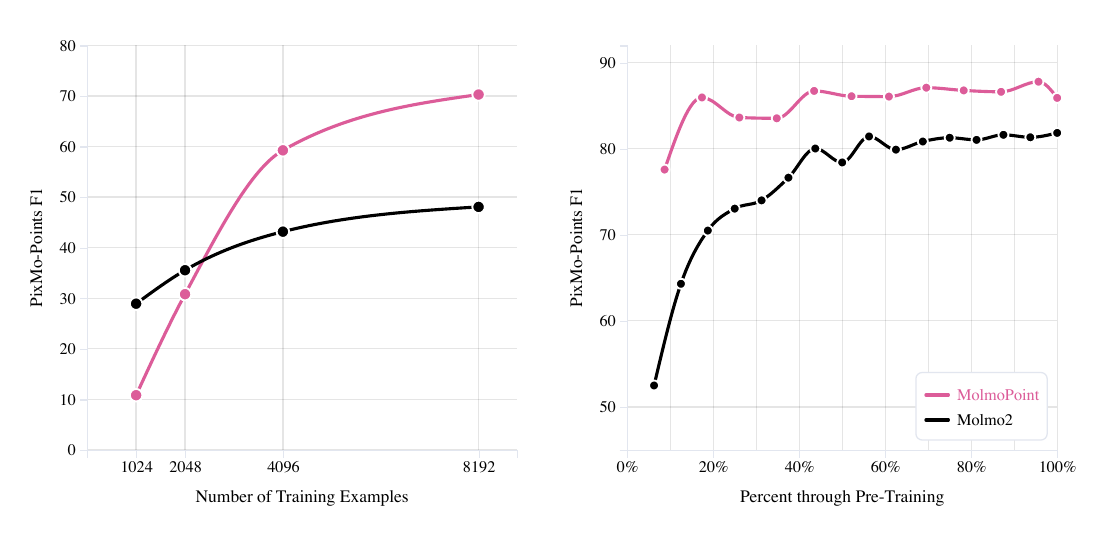}


\label{fig:data_efficiency}
\end{figure}

Figure~\ref{fig:data_efficiency} (left) shows performance when fine-tuning a base captioning model on a small number of pointing examples. Models were tuned for 6 epochs to ensure they were fully saturated. \model{} initially performs worse, likely due to needing to learn new parameters from scratch, but quickly improves to a 20 point gain when using 8192 examples (the full pointing dataset has close to half a million examples). 
Figure~\ref{fig:data_efficiency} (right) shows MolmoPoint reaches peak performance faster during pre-training. Both of these results show that grounding tokens make pointing easier and more efficient to learn than text coordinates. 

\subsection{Qualitative results}
\begin{figure}[!t]
    \centering
    \includegraphics[width=\linewidth]{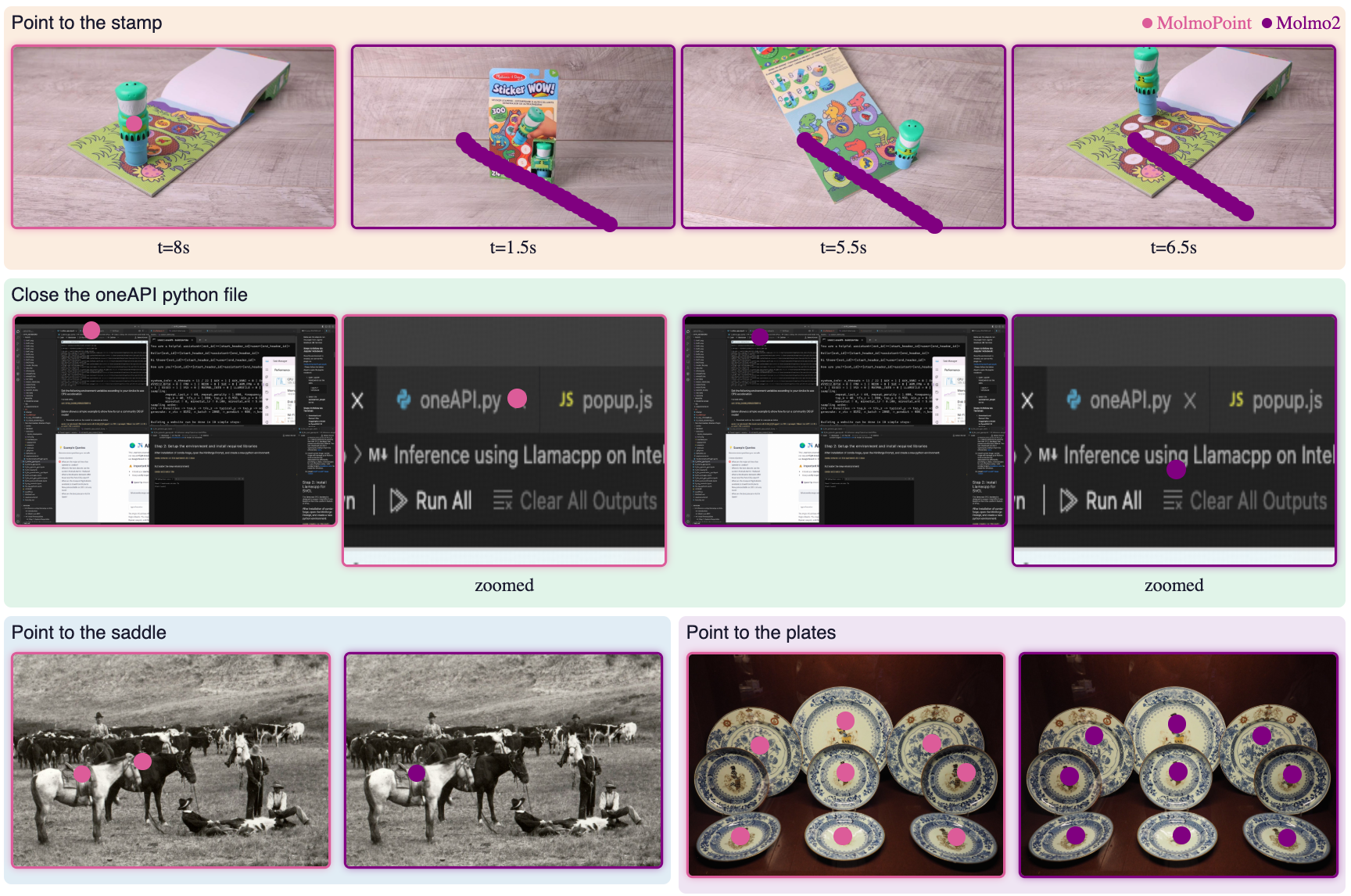}
    \caption{\textbf{Qualitative examples}. Top, Molmo2 generates lines of incorrect points in multiple video frames. Middle, Molmo2 is unable to localize the ``x'' exactly (zoomed images are close-ups of the same point). Bottom left, \model{} finds the second, partly occluded saddle. Bottom right, \model{} misses one of the plates. The middle row shows results from \modelImg{} and Molmo2-GUI-8B, and the others show \model{} and Molmo2-8B.}
    \label{fig:qual}
\end{figure}

Despite being trained on the same data, we observe significant qualitative differences between \model{} and Molmo2-8B.
\model{} is less likely to produce degenerate output on videos, is better at finding small objects, and can be more precise when pointing. However, we observe that it occasionally produces off-by-one errors when counting high-frequency objects. Figure~\ref{fig:qual} demonstrates qualitative examples to compare Molmo2 and Molmo2-P in video pointing, high-resolution GUI grounding, and multi-object pointing.

\section{Conclusion}
We have shown that using grounding tokens significantly improves pointing across multiple domains.
The improvements in sample efficiency and training speed show this method would be especially helpful in low-resource settings.
Future work could extend this approach to include other modalities, such as pointing to text tokens to highlight important parts of the text or pointing to audio tokens to reference a sound.

\section*{Acknowledgements}
This work would not be possible without the support of our colleagues at Ai2.

\begin{itemize}
    \item We thank David Albright, Cailin Brashear, Crystal Nam, Kyle Wiggers, and Will Smith for their important work for the \model public release.
    \item We thank other members of the PRIOR team for providing advice and feedback on various aspects of \model{}.
    \item We thank the Prolific team for their support and our annotators on Prolific for providing us with high-quality data that is crucial to \model{}. 
\end{itemize}

\noindent This material is based upon work supported by the National Science Foundation under Award No. 2413244.

\clearpage

\bibliographystyle{abbrvnat}
\bibliography{main, molmov1}

\clearpage

\section*{Appendix}

This appendix includes the following sections:
\begin{itemize} 
\itemsep0em 
    \item \S\ref{sect:training} - Training Details
    \item \S\ref{sect:video_point_human_eval} - VideoPoint Human Eval
    \item \S\ref{sect:multitask_results} - Multi-Task Results
    \item \S\ref{sect:guidata} - \guidata{} Details
    \item \S\ref{sect:trackdata} - \trackdata{} Details
\end{itemize}

\section{Training Details}
\label{sect:training}
Our training pipeline largely follows Molmo2, with a few differences to account for the hardware we have available, improve efficiency, and take advantage of the increased learning speed when using grounding tokens. We also expand the tracking data from Molmo2 to improve tracking robustness. We discuss the training changes below, see Section~\ref{sect:trackdata} for details on the tracking data. 

\paragraph{Pre-training.}
We improve packing efficiency by allowing up to 16 images per input sequence. We reduce the total number of training steps from 32k to 23k to keep the number of examples seen unchanged. We use a learning rate of $1\mathrm{e}{-4}$ with a warmup of 200 for the pointing parameters.

\begin{table}[]
    \caption{\textbf{Updated top-level sampling rates}. For \model{} we slightly reduce the sampling rates on pointing tasks since we observe faster convergence of these tasks. Each of these categories consists of its own sub-mixture of datasets. We leave the sampling ratios within those submixtures unchanged. See \cite{molmo2} for the full list.}
    \label{tab:sampling_rates}
    \centering
    \small
    \begin{tabular}{lcc}
         \textbf{Dataset Group} & \textbf{\model{}} & \textbf{Molmo2} \\
         \toprule
         Captioning/Long QA & 15.0 & 13.6 \\
         Image QA & 25.0 & 22.7 \\
         Video QA & 20.0 & 18.2 \\
         Image Pointing & 7.0 & 9.1 \\
         Video Pointing & 11.0 & 13.6 \\
         Video Tracking & 12.0 & 13.6 \\
         NLP & 10.0 & 9.1 \\
    \end{tabular}
\end{table}

\paragraph{SFT.}
We slightly tweak the mixture rates of Molmo2 because preliminary experiments showed that grounding tasks converged significantly faster when using grounding tokens. We do this by adjusting the top-level sampling rates used in the Molmo2 training pipeline; see Table~\ref{tab:sampling_rates}. 
We also train for 22k steps with a batch size of 160 instead of 30k steps with a batch size of 128, which reduces the number of examples seen by about 8.3\%. 
We still use a learning rate of $1\mathrm{e}{-4}$ for the pointing parameters.

\paragraph{Long-context SFT.}
We train with a batch size of 160 instead of 128, and we train with a maximum of 384 frames and 16384 tokens per training example, following Molmo2.

\paragraph{Specialized models.}
Both \modelVid{} and \modelImg{} used the same optimizer settings as the SFT model. We train \modelImg{} for 2000 steps with a batch size of 128 and a maximum of 48 crops per image on \guidata{}.
We train \modelVid{} using the same pre-training stage, and then training on just the Video Point dataset group from Molmo2 for 6000 steps and a batch size of 64, and then for another 800 steps with a maximum of 384 frames.

\paragraph{Training times}.
Full training times are shown in Table~\ref{tab:training_times}. All training was done on B200 GPUs. Following Molmo2, we use PyTorch with Torch's Fully Sharded Data Parallel (FSDP) 2~\cite{zhao2023pytorch} for distributed training and the Automatic Mixed Precision (AMP) module\footnote{\url{https://docs.pytorch.org/docs/stable/amp.html}} for mixed precision training.
We do not use sequence-parallelism since we found that a long-context model can fit on the B200 GPUs with just FSDP.

\begin{table}[]
\small
    \caption{\textbf{Training times}. Columns show the model, the training phase, gpu counts, training hours, total GPU hours, batch size, the estimated number of multi-modal inputs that were seen (separately counting inputs that were packed together), and the estimated number of training annotations that were seen (separately counting annotations that were merged into message trees).}
    \label{tab:training_times}
    \centering
    \begin{tabular}{lccccccccc}
         \textbf{Model} & \textbf{phase} & \textbf{Hardware} & \textbf{GPUs} & \textbf{hours} & \textbf{GPU hr.} & \textbf{batch} & \textbf{steps} & \textbf{mm.} & \textbf{annotations} \\
         \toprule
\model{} & Pretrain & B200 & 32 & 10.3 & 330 & 128 & 23k & 4.6m & 14m \\
\model{} & SFT & B200 & 80 & 86.4 & 6.9k & 160 & 22k & 13m & 49m \\
\model{} & SFT-LC & B200 & 80 & 36.2 & 2.9k & 160 & 2k & 2.2m & 7.5m \\
\modelImg{} & SFT & B200 & 32 & 9.4 & 300 & 128 & 2k & 360k & 11m \\
\modelVid{} & Pretrain & H100 & 32 & 10.1 & 323 & 128 & 23k & 3.5m & 6.6m \\
\modelVid{} & SFT & B200 & 32 & 13.7 & 437 & 64 & 6k & 1.4m & 5.1m \\
\modelVid{} & SFT-LC & B200 & 32 & 12.5 & 400 & 64 & 800 & 170k & 270k \\
    \end{tabular}
\end{table}

\section{VideoPoint Human Evaluation}
\label{sect:video_point_human_eval}
We perform an internal human preference evaluation with two annotators (two of the authors) on the video pointing outputs of two models -- \model{} and the text coordinate baseline. We first manually design the test set by selecting 271 challenging video-query pairs from Molmo2-VideoCountEval~\cite{molmo2} and writing 199 new queries for the videos in Molmo2-VideoCaptionEval, resulting in a total of 470 examples in this human preference evaluation dataset. We develop a simple annotation interface to collect pair-wise preferences on this evaluation dataset following standard practice~\cite{chiang2024chatbot}, where the ordering of models a and b is randomized, and four choices are allowed: model a is better; model b is better; both are good; or both are bad. After collecting all preferences, we then compute the win rate of \model{} against the baseline, excluding ties (i.e., both good or both bad).

\newcommand{\mycell}[2]{%
  \rotatebox{90}{%
    \parbox{2.5cm}{%
      \setlength{\baselineskip}{0.5em}%
      \textbf{\scriptsize{#1}}\\
      \footnotesize{\textcolor{gray}{#2}}%
    }%
  }%
}

\begin{table*}[!t]
    \centering
    \caption{\textbf{Video benchmark results} for a range of proprietary APIs, open-weight baselines, fully-open baselines, Molmo2, and \model{} across video understanding, captioning, and counting benchmarks. The best-performing open-weight model is in \textbf{bold}, and the second-best is \underline{underlined}. For \model{}, we report PerceptionTest val since the test server is no longer operational.}
    \label{tab:video_benchmark_results}
    
    \setlength{\tabcolsep}{2.5pt}
    \resizebox{\textwidth}{!}{
    \begin{tabular}{@{}l>{\columncolor{tableyellow!50}}c
    >{\columncolor{tableyellow!50}}c
    >{\columncolor{tableyellow!50}}c
    >{\columncolor{tableyellow!50}}c
    >{\columncolor{tableyellow!50}}c
    >{\columncolor{tableyellow!50}}c
    >{\columncolor{tablegreen!10}}c
    >{\columncolor{tablegreen!10}}c
    >{\columncolor{tablegreen!10}}c
    >{\columncolor{tablegreen!10}}c
    >{\columncolor{tablegreen!10}}c
    >{\columncolor{tablegreen!10}}c
    >{\columncolor{tablegreen!10}}ccc
    >{\columncolor{tableyellow!50}}c
    >{\columncolor{tablegreen!10}}c
    >{\columncolor{tableyellow!50}}c@{}}
        \textbf{Model} & 
        \mycell{NextQA}{test~\cite{nextqa}}
        & \mycell{PerceptionTest}{test~\cite{perception_test}}
        & \mycell{MVBench}{test~\cite{mvbench}}
        & \mycell{Tomato}{test~\cite{tomato}}
        & \mycell{MotionBench}{val~\cite{motionbench}}
        & \mycell{TempCompass}{test MCQ~\cite{tempcompass}}
        & \mycell{Video-MME}{test~\cite{videomme}}
        & \mycell{Video-MME-Sub}{test~\cite{videomme}}
        & \mycell{LongVideoBench}{val~\cite{longvideobench}}
        & \mycell{MLVU}{test MCQ~\cite{mlvu}}
        & \mycell{LVBench}{test~\cite{lvbench}}
        & \mycell{VideoEvalPro}{test~\cite{videoevalpro}}
        & \mycell{Ego Schema}{test~\cite{egoschema}}
        & \mycell{Molmo2 Caption}{test F1 Score}
        & \mycell{Molmo2 Count}{val accuracy}
        & \newcell{Short QA avg.}
        & \newcell{Long QA avg.}
        & \newcell{Average}\\
        \midrule
        
        \apionlyheader{12} \\

        GPT-5~\cite{gpt5}              & 86.3 & 79.4 & 74.1 & 53.0 & 65.4 & 80.4 & 83.3 & 86.9 & 72.6 & 77.7 & 65.2 & 68.8 & 75.6 & 50.1 & 35.8 & 73.1 & 76.3 & 70.6\\
        GPT-5 mini~\cite{gpt5}         & 83.2 & 72.0 & 66.5 & 44.1 & 59.9 & 74.9 & 77.3 & 82.3 & 69.7 & 69.1 & 54.7 & 60.1 & 70.9 & 56.6 & 29.8 & 66.8 & 69.8 & 65.0\\
        Gemini 3 Pro~\cite{gemini3} & 84.3 & 77.6 & 70.4 & 48.3 & 62.6 & 82.8 & 88.6 & 87.5 & 75.9 & 75.7 & 77.0 & 78.0 & 68.9 & 36.0 &37.1 & 71.0	& 78.8 & 70.0\\
        Gemini 2.5 Pro~\cite{comanici2025gemini}     & 85.3 & 78.4 & 70.6 & 48.6 & 62.0 & 81.9 & 87.8 & 87.8 & 76.8 & 81.5 & 75.7 & 78.4 & 72.2 & 42.1 & 35.8 & 71.1 & 80.4 & 71.2\\
        Gemini 2.5 Flash~\cite{comanici2025gemini}   & 81.8 & 74.7 & 67.0 & 39.1 & 59.3 & 80.2 & 84.2 & 84.2 & 73.1 & 75.1 & 64.9 & 69.6 & 70.2 & 46.0 & 31.9 & 67.0 & 74.5 & 66.7\\
        Claude Sonnet 4.5~\cite{anthropic2025sonnet}  & 79.2 & 64.3 & 62.1 & 39.6 & 58.5 & 72.8 & 74.2 & 80.5 & 65.1 & 64.0 & 50.5 & 50.5 & 73.1 & 26.0 & 27.2 & 62.8 & 66.4 & 59.6\\
        \midrule
        \openweightsheader{12} \\

        InternVL3.5-4B~\cite{wang2025internvl3}     & 80.3 & 68.1 & 71.2 & 26.8 & 56.5 & 68.8 & 65.4 & 68.6 & 60.8 & 52.0 & 43.2 & 46.5 & 58.9 & 7.7 & 26.3  & 62.0 & 56.5 & 53.4\\

        InternVL3.5-8B~\cite{wang2025internvl3}     & 81.7 & 72.7 & 72.1 & 24.6 & 56.6 & 70.3 & 66.0 & 68.6 & 62.1 & 53.2 & 43.4 & 48.1 & 58.6 & 7.8 & 26.1  & 63.0 & 57.1 &  54.1\\

        Qwen3-VL-4B~\cite{qwen3technicalreport}         & 81.4 & 70.7 & 68.9 & 31.8 & 58.6 & 70.8 & 69.3 & 74.0 & 62.8 & 58.4 & \underline{56.2} & 49.8 & 68.4 & 25.2 & 25.3 & 63.7 & 62.7 & 58.1\\

        Qwen3-VL-8B~\cite{qwen3technicalreport}         & 83.4 & 72.7 & 68.7 & 35.7 & 56.9 & 74.3 & 71.4 & 75.2 & 62.4 & 57.6 & \textbf{58.0} & 50.3 & \underline{69.8} & 26.7 & 29.6 & 65.3 & {63.5} & 59.5\\

        Keye-VL-1.5-8B~\cite{yang2025kwai}        & 75.8 & 64.2 & 56.9 & 33.0 & 55.1 & \textbf{75.5} & \textbf{73.0} & \textbf{76.2} & 66.0 & 53.8 & 42.8 & 54.9 & 56.3 & 25.4 & 27.2 & 60.1 & 60.4 & 55.7\\

        GLM-4.1V-9B~\cite{glmv}          & 81.3 & 74.2 & 68.4 & 30.0 & 59.0 & 72.3 & 68.2 & 75.6 & 65.7 & 56.6 & 44.0 & 51.1 & 62.6 & 18.4 & 26.6 & 64.2 & 60.5 & 56.9\\

        MiniCPM-V-4.5-8B~\cite{yu2025minicpmv45cookingefficient}      & 78.8 & 70.9 & 60.5 & 29.8 & 59.7 & 72.7 & 67.9 & 73.5 & 63.9 & \underline{60.6} & 50.4 & 54.9 & 49.6 & 29.3 & 26.3 & 62.1 & 60.1 & 56.6\\

        Eagle2.5-8B~\cite{eagle2_5}             & 85.0 & 81.0 & 74.8 & 31.0 & 55.7 & \underline{74.4} & \underline{72.4} & 75.7 & 66.4 & 60.4 & 50.9 & 58.6 & \textbf{72.2} & 22.8 & 28.9 & 67.0 & \textbf{65.2} & 60.7\\

        \midrule

        \fullyopenheader{12} \\
        PLM-3B~\cite{cho2025PerceptionLM}             &83.4 & 79.3 & 74.7 & 30.9 & 60.4 & 69.3 & 54.9 & 59.4 & 57.9 & 48.4 & 40.4 & 46.2 & 66.9 & 12.3 & 24.4 & 66.3 & 53.5 & 53.9\\

        PLM-8B~\cite{cho2025PerceptionLM}            & 84.1 & \textbf{82.7} & \textbf{77.1} & 33.2 & 61.4 & 72.7 & 58.3 & 65.4 & 56.9 & 52.6 & 44.5 & 47.2 & 68.8 & 10.9 & 26.6 & 68.5 & 56.2 & 56.2\\
        
        LLaVA-Video-7B~\cite{llava_video}     & 83.2 & 68.8 & 58.6 & 24.9 & 54.2 & 66.6 & 63.3 & 69.7 & 58.2 & 52.8 & 44.2 & 47.8 & 57.3 & 19.9 & 21.4 & 59.4 & 56.2& 52.7\\

        VideoChat-Flash-7B~\cite{li2024videochat}    & \underline{85.5} & 76.5 & 74.0 & 32.5 & 60.6 & 69.4 & 65.3 & 69.7 & 64.7 & 56.0 & 48.2 & 51.2 & 51.3 & 14.8 & 21.6 & 66.4 & 58.1 & 56.1\\

        \textcolor{molmocolor}{Molmo2-4B}     & \underline{85.5} & 81.3 & 75.1 & \textbf{39.8} & \underline{61.6} & 72.8 & 69.6 & 75.7 & \textbf{68.0} & \textbf{63.0} & 53.9 & \underline{59.9} & 61.2 & 39.9 & \underline{34.3} & \underline{69.3} & \underline{64.5} & \underline{62.8}\\

        \textcolor{molmocolor}{Molmo2-8B}      & \textbf{86.2} & \underline{82.1} & \underline{75.9} & \underline{39.6} & \textbf{62.2} & 73.4 & 69.9 & \underline{75.8} & \underline{67.5} & 60.2 & 52.8 & \textbf{60.4} & 62.0 & \textbf{43.2} & \textbf{35.5} & \textbf{69.9}	&64.1 & \textbf{63.1}\\

        \textcolor{molmocolor}{Molmo2-O-7B}          & 84.3 & 79.6 & 74.8 & 36.2 & 60.6 & 73.0 & 64.9 & 69.2 &	63.7 & 55.2 & 49.6 & 55.1 & 56.8 & \underline{40.1} & 33.2 & 68.1	& 59.2 & 59.7\\

        \midrule
        \modelheader{7} \\
        \textcolor{molmocolor}{\model{}} & 86.3 & 82.3\rlap{$^*$} & 75.2 & 37.3 & 61.4 & 74.4 & 69.6 & 76.9 & 67.3 & 62.4 & 52.5 & 58.0 & 63.0 & 40.3  & 35.6 & 69.5 & 64.2 & 63.5
    \end{tabular}
    }%
\end{table*}

\section{Multi-Task Results}
\label{sect:multitask_results}
We present the full set of video results in Table~\ref{tab:video_benchmark_results}.
Comparing Molmo2 and \model{} on video QA, we see mixed outcomes on the long-video QA benchmarks (MLVU (+2.2), Video-MME-SUB (+1.1), VideoEvalPro (-2.4)) and on short-video temporal understanding benchmarks (TempCompass (+1.0), Tomato (-2.3)), with results being about the same elsewhere. We do not see a clear pattern in these results and conclude that the models perform very similarly.

The full set of image benchmarks is shown in Table~\ref{tab:image_benchmark_results}.
Compared to Molmo2, we see notable gains on some high-res/OCR-tasks (InfoQa (+2.9), Ai2D (+0.8), ChartQA (+0.8)) and a slight drop in counting benchmarks (CountBench (-0.8), PixMoCount (-0.9)). 
The consistent gain on high-res/OCR tasks suggests that grounding tokens improved transfer from the pointing training datasets to these tasks, which continues the theme of improving high-res OCR-heavy tasks that we observe with GUI pointing. We attribute the slight drop in counting to the occasional off-by-one errors we observe when counting high-frequency objects (see Figure~\ref{fig:qual} in the main paper). 

There is a significant drop in video captioning performance. However, when we check image-captioning performance, we see the opposite: 54.47 for \model{} vs 53.62 for Molmo2, using the F1 captioning metric from ~\cite{molmov1}. This again suggests that the way grounding tokens affect video and image performance differs.

\renewcommand{\mycell}[2]{%
  \rotatebox{90}{%
    \parbox{2.0cm}{%
      \setlength{\baselineskip}{0.5em}%
      \textbf{\scriptsize{#1}}\\
      \footnotesize{\textcolor{gray}{#2}}%
    }%
  }%
}

\renewcommand{\newcell}[1]{%
  \rotatebox{90}{%
    \parbox{2.0cm}{%
      \setlength{\baselineskip}{0.5em}%
      \textbf{\scriptsize{#1}}
    }%
  }%
}

\begin{table*}[!t]
    \caption{\textbf{Image benchmark results} for a range of proprietary APIs, open-weight baselines, and \model{} across a range of image understanding and counting benchmarks. The result of the best-performing open-weight model is in \textbf{bold}, and the second best is \underline{underlined}.}
    \label{tab:image_benchmark_results}
    
    \centering
    \resizebox{\textwidth}{!}{
    \begin{tabular}{@{}l>{\columncolor{tableyellow!50}}c
    >{\columncolor{tableyellow!50}}c
    >{\columncolor{tableyellow!50}}c
    >{\columncolor{tableyellow!50}}c
    >{\columncolor{tableyellow!50}}c
    >{\columncolor{tableyellow!50}}c
    >{\columncolor{tableyellow!50}}c
    >{\columncolor{tableyellow!50}}c
    >{\columncolor{tableyellow!50}}c
    >{\columncolor{tableyellow!50}}c
    >{\columncolor{tableyellow!50}}c
    >{\columncolor{tablegreen!10}}c
    >{\columncolor{tablegreen!10}}c
    >{\columncolor{tablegreen!10}}c
    >{\columncolor{tableyellow!50}}c
    >{\columncolor{tablegreen!10}}c
    >{\columncolor{tableblue!10}}c}
        \textbf{Model} & 
        \mycell{AI2D}{test~\cite{ai2_diagram}}
        & \mycell{ChartQA}{test~\cite{chartqa}}
        & \mycell{DocVQA}{test~\cite{mathew2021docvqa}}
        & \mycell{InfoQA}{test~\cite{infoqa}}
        & \mycell{TextVQA}{val~\cite{textqa}}
        & \mycell{VQA v2.0}{val ~\cite{goyal2017making}}
        & \mycell{RWQA}{\cite{realworldqa}}
        & \mycell{MMMU}{val~\cite{yue2024mmmu}}
        & \mycell{MathVista}{testmini~\cite{lu2024mathvista}}
        & \mycell{CountBench}{~\cite{beyer2024paligemma}}
        & \mycell{PixMoCount}{test \cite{molmov1}}
        & \mycell{MuirBench}{\cite{wang2024muirbench}}
        & \mycell{MMIU}{\cite{meng2024mmiumultimodalmultiimageunderstanding}}
        & \mycell{Blink}{val \cite{fu2024blink}}
        & \newcell{Img QA avg.}
        & \newcell{MultiImg QA avg.}
        & \newcell{Average}\\
        \midrule
        
        \multicolumn{16}{@{}l}{\textbf{\textit{API call only}}} \\
        GPT-5~\cite{gpt5}        & 89.5 & 83.8 & 88.9 & 83.0 & 78.7 & 79.7 & 80.8 & 81.8 & 82.7 & 90.8 & 67.2  & 78.6 & 71.0 & 66.5 & 82.5 & 72.1 & 80.2 \\

        GPT-5 mini~\cite{gpt5}  & 86.7 & 82.1 & 86.7 & 82.2 & 79.1 & 72.1 & 77.0 & 78.7 & 79.2 & 87.1 & 74.4 & 71.4 & 64.5 & 68.7 & 80.5 & 68.2 & 77.8  \\

        Gemini 2.5 Pro~\cite{comanici2025gemini} & 94.3 & 77.8 & 91.5 & 82.0 & 70.3 & 67.1 & 77.4 & 79.6 & 84.6 & 90.8 & 73.8 & 74.5 & 68.9 & 73.7 & 80.8 & 72.4 & 79.0 \\

        Gemini 2.5 Flash~\cite{comanici2025gemini}   & 95.9 & 76.8 & 91.1 & 80.9 & 73.0 & 69.4 & 74.5 & 79.0 & 81.2 & 86.7 & 63.9 & 73.5 & 61.2 & 70.2 & 79.3 & 68.3 & 76.9 \\

        Claude Sonnet 4.5~\cite{anthropic2025sonnet}  & 91.5 & 80.2 & 91.7 & 65.9 & 67.2 & 77.0 & 61.1 & 77.8 & 73.1 & 87.3 & 58.3 & 59.6 & 54.1 & 64.8 & 75.6 & 59.5 & 72.1 \\

        \midrule
        
        \multicolumn{16}{@{}l}{\textbf{\textit{Open weights only}}} \\

        InternVL3.5-4B~\cite{wang2025internvl3} & 82.6 & 86.0 & 92.4 & 78.0 & 77.9 & 78.1 & 66.3 & 66.6 & 77.1 & 82.2 & 62.4 & 53.1 & 49.2 & 58.1 & 77.2 & 53.5 & 72.1 \\

        InternVL3.5-8B~\cite{wang2025internvl3} & 84.0 & 86.7 & 92.3 & 79.1 & 78.2 & 79.5 & 67.5 & \textbf{73.4} & 78.4 & 79.6 & 61.9 & 55.8 & 49.4 & 59.5 & 78.2 & 54.9 & 73.2 \\

        Qwen3-VL-4B~\cite{qwen3technicalreport} & 84.1 & 85.0 & \underline{95.3} & 80.3 & 81.0 & 81.7 & 70.9 & 67.4 & 73.7 & 85.5 & 58.0 & 63.8 & 43.2 & \underline{65.8} & 78.4 & 57.6 & 74.0 \\

        Qwen3-VL-8B~\cite{qwen3technicalreport}      & 85.7 & 85.2 & \textbf{96.1} & \textbf{83.1} & 82.8 & 82.3 & 71.5 & 69.6 & 77.2 & 90.4 & 65.0 & \underline{64.4} & 35.3 & \textbf{69.1} & 80.8 & 56.3 & 75.6 \\

        Keye-VL-1.5-8B~\cite{yang2025kwai}    & 89.5 & 85.0 & 93.4 & 74.9 & 81.5 & 79.3 & 73.5 & \underline{71.4} & \textbf{81.2} & 81.6 & 57.4 & 51.2 & 50.3 & 54.9 & 79.0 & 52.1 & 73.2 \\

        GLM-4.1V-9B~\cite{glmv}      & 87.9 & 70.0 & 93.3 & 80.3 & 79.6 & 68.3 & 70.7 & 68.0 & \underline{80.7} & 88.0 & 60.7 & \textbf{74.7} & \textbf{62.4} & 65.1 & 77.0 & \textbf{67.4} & 75.0 \\

        MiniCPM-V-4.5-8B~\cite{yu2025minicpmv45cookingefficient}  & 86.5 & \underline{87.4} & 94.7 & 73.4 & 82.2 & 64.1 & 72.1 & 67.7 & 79.9 & 83.9 & 62.8 & 53.3 & 46.5 & 42.0 & 77.7 & 47.3 & 71.2 \\

        Eagle2.5-8B~\cite{eagle2_5}    & 84.5 & \textbf{87.5} & 94.1 & 80.4 & 83.7 & 82.4 & 76.7 & 55.8 & 67.8 & 90.2 & 66.9 & 61.8 & 48.4 & 45.8 & 79.1 & 52.0 & 73.3 \\

        \midrule
        
        \fullyopenheader{12} \\
        PLM-3B~\cite{cho2025PerceptionLM} &  90.9 & 84.3 & 93.8 & 74.6 & 84.3 & 84.4 & 72.4 & 41.2 & 59.1 & 87.1 & 63.0 & 25.7 & 40.6 & 55.4 & 75.9 & 40.6 & 68.3 \\

        PLM-8B~\cite{cho2025PerceptionLM}& 92.7 & 85.5 & 94.6 & 80.0 & \textbf{86.5} & 85.6 & 75.0 & 46.1 & 59.9 & 91.8 & 68.0 & 23.5 & 27.4 & 56.0 & 78.7 & 35.7 & 69.5 \\

        \textcolor{molmocolor}{Molmo2-4B} & 95.6 & 86.1 & 87.8 & 78.6 & 85.0 & 86.6 & 75.4 & 50.9 & 56.7 & \underline{93.9} & 88.1 & 60.5 & \underline{55.5} & 57.5 & 80.4 & \underline{57.8} & 75.6 \\
        \textcolor{molmocolor}{Molmo2-8B} & \underline{95.8} & 86.0 & 93.2 & 80.1 & 85.7 & \underline{87.0} & \textbf{77.6} & 53.0 & 58.9 & 93.7 & \underline{88.5} & 63.7 & 54.2 & 51.3 & \underline{81.7} & 56.4 & \underline{76.3} \\
        \textcolor{molmocolor}{\model-O-7B} & 93.7 & 84.9 & 90.4 & 77.9 & 84.7 & 86.6 & 73.6 & 45.8 & 54.2 & \textbf{95.1} & \textbf{88.9} & 58.4 & 51.7 & 50.5 & 79.7 & 53.5 & 74.1 \\
        \midrule
        \modelheader{12} \\
        \model{} & \textbf{96.4} & 86.8 & 93.8 & \underline{83.0} & \underline{86.0} & \textbf{87.2} & \underline{77.4} & 53.7 & 59.4 & 92.9 & 87.6 & 62.5 & 54.6 & 50.8 & \textbf{82.2} & 56.0 & \textbf{76.6}
    \end{tabular}
    }
\end{table*}

\section{\guidata{} Details}
\label{sect:guidata}

We show the data generation pipeline of \guidata{} in Figure~\ref{fig:molmop_syn}. 
The input to the pipeline is a natural language query, e.g., ``a screenshot of AutoCAD'', which will be paired with a randomly selected persona from PersonaHub~\cite{ge2024scaling} (e.g., \textit{a Sci-fi novelist})to diversify its content and style. 

We systematically construct a comprehensive list of queries by considering screenshot types (desktop, web, mobile), task domains (different websites, Apps, software), platforms (Windows, macOS, iOS, Android, etc.), aspect ratios (4:3, 16:9, etc.), resolutions (720p, 1080p, 4K, etc.), and stages during a task (early, middle, end).
We randomly sample and combine those fields to construct inputs that span a broad range of scenarios in the digital world.

We feed the query into our prompt template, and an LLM outputs the corresponding HTML code to render the screenshot. 
We run customized JavaScript on the HTML code to extract the bounding boxes for all visible elements in the screenshot.
Each bounding box contains the synthetic label from its naming attributes in HTML, the corresponding lines of code for this element, the (x, y) center, and the (width, height) of the box.
We feed this information back to the LLM to annotate each element with a natural language name (e.g., ``Measure Button'') and 5 different intents that a real user might ask when interacting with this element.

\begin{figure}[!t]
    \centering
    \includegraphics[width=0.99\linewidth]{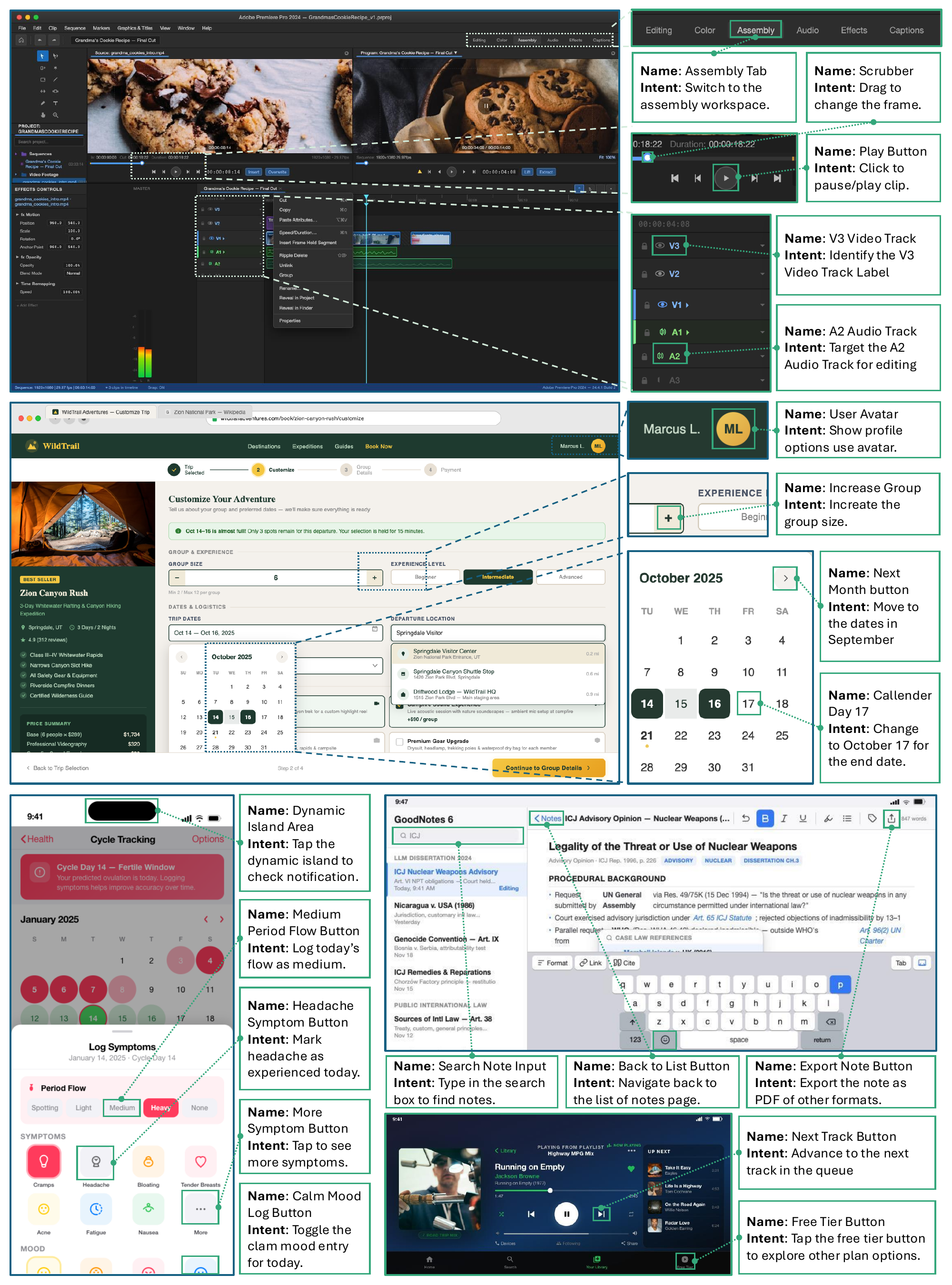}
    \caption{\textbf{Qualitative examples of \guidata{}.} We demonstrate GUI grounding examples for desktop, web, and mobile screenshots.}
    \label{fig:gui_example}
\end{figure}

We use claude-sonnet-4.6 as our coding LLM to generate \guidata{}, which costs about \$0.2 per example, with an average of 54 pointing annotations.
Figure \ref{fig:gui_example} demonstrates the qualitative examples from \guidata{}.

\section{\trackdata{} Details}
\label{sect:trackdata}
Here, we detail the data generation pipeline for \trackdata{},
which comprises two complementary  data sources: (1) \trackreal{}, human-annotated tracks covering a broad range of videos and object categories, and (2) \tracksynth{}, synthetically generated object tracks featuring complex occlusion patterns and motion dynamics. 

\subsection{\trackreal{}: Human Annotated Tracks}
To extend tracking annotations beyond existing datasets, we develop a human-in-the-loop pipeline for annotating object tracks in real videos. Annotating point tracks from scratch is both costly and difficult to quality-control; for multi-object scenes, annotators must identify unique instances, track them simultaneously, and handle highly variable workloads across videos. However, if the objects of interest and their count are known in advance, the task simplifies to tracking a single designated object at a time.

Specifically, we leverage the Molmo2-VideoPoint data, which already provides distinct identities for each text query, and extend single-frame points into full tracks. Figure~\ref{fig:molmotrack_real} illustrates the overall data generation pipeline. Each annotation task provides the annotator with the video, the input point for the object of interest, and all other annotated points for context. Annotators track one object at a time while viewing the surrounding points, which helps prevent duplicate tracks for the same instance, particularly in ambiguous cases involving shot changes or visually similar objects. By reducing the cognitive load to single-object tracking, annotators no longer need to jointly identify and track multiple objects. In the end, our annotation consists of 13K videos with 17K text queries and an average of 6.7 unique objects per video, accompanied by diverse tracks with re-identification in shot changes, part-level objects, and complex referring expressions.

\begin{figure}[!t]
    \centering
    \includegraphics[width=0.99\linewidth]{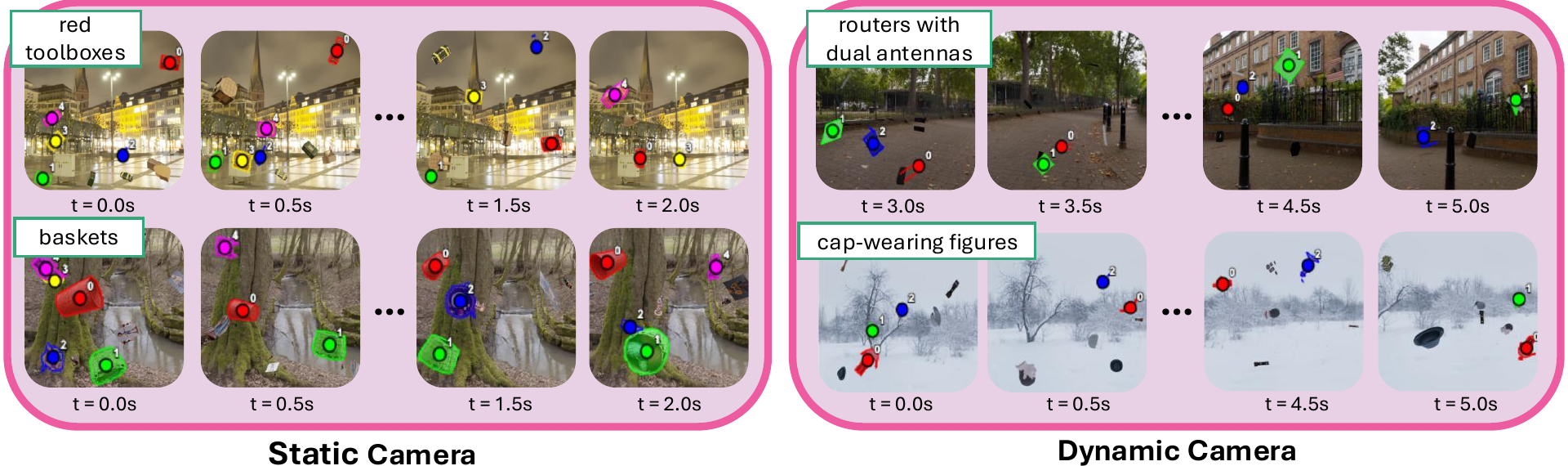}
    \caption{\textbf{Examples from \tracksynth{} data.} Object tracks, text queries, and videos are generated synthetically with static and dynamic camera viewpoints.}
    \label{fig:molmotrack_syn}
\end{figure}

\subsection{\tracksynth{}: Synthetic Object Tracks and Videos}

We generate synthetic multi-object tracking videos in Blender using two pipelines: a static-camera pipeline and a moving-camera pipeline (see Figure~\ref{fig:molmotrack_syn} for examples). In both cases, each video is rendered with configurable duration, frame rate, resolution, and physically based rendering settings, and the pipeline outputs RGB frames together with per-instance binary masks derived from Blender’s object-index pass. The static-camera version uses a fixed camera pose and samples object trajectories relative to a camera-aware frustum, while the moving-camera version extends this setup with a smoothly animated camera and frame-dependent waypoint planning so that visibility constraints are enforced with respect to the camera motion at each waypoint. After generation with diverse motion patterns, the frames are then encoded uniformly into videos as 6 fps. 

In both pipelines, 3D assets are sampled from TexVerse~\cite{zhang2025texverse} after automatic caption-based filtering with GPT. The filter keeps only independently trackable objects and removes scenes, backgrounds, abstract assets, oversized context assets, and collections of unrelated objects, while also assigning noun-only semantic categories to retained assets. For each video, we first build category-specific pools from the filtered assets, randomly choose 1--3 categories, allocate the requested number of objects across these categories, and then sample distinct assets accordingly. This category-based sampling encourages semantic diversity within each sequence while avoiding duplicate object identities.

After selection, assets are imported, merged if needed, normalized to a target scale, centered, and placed on the ground plane. The scene is rendered with randomized lighting and camera parameters. Object motion is generated by sampling waypoint-based trajectories under camera-aware visibility constraints, so objects can be forced to stay visible or move off-screen for selected time spans. The moving-camera pipeline further animates the camera along a smooth trajectory and recomputes visibility constraints with respect to the camera pose over time.
For each sequence, we render RGB frames and per-instance segmentation masks, then convert the masks into frame-wise tracking annotations with consistent object identities across time.

After synthesizing the videos, we automatically generate language queries from the selected object captions associated with each sequence. For every video, we collect the source captions of all sampled objects and prompt GPT to produce 0--3 referring queries, where each query must describe a group containing at least two objects. The query generator is constrained to produce concise group-level noun phrases rather than enumerations, avoid explicit counts, and avoid references to object parts, scenes, camera state, or temporal events. It may use either shared fine-grained types (e.g., the toolboxes) or higher-level categories (e.g., the vehicles) when grouping objects. The resulting queries are then matched back to object identities and stored together with the video path and frame-wise tracking annotations, yielding language-conditioned multi-object tracking examples.

After generating the segmentation masks, we extract the center of the largest connected component to obtain the point tracks. Overall, we have 76k unique queries for 25k videos as our training data, with an average of 3.3 unique objects per video.

\end{document}